\documentclass{article}

\usepackage{microtype}
\usepackage{graphicx}
\usepackage{booktabs} %

\usepackage{hyperref}

\usepackage{caption}
\usepackage{subcaption}

\usepackage[preprint]{icml2025}

\usepackage{amsmath}
\usepackage{amssymb}
\usepackage{mathtools}
\usepackage{amsthm}

\usepackage[capitalize,noabbrev]{cleveref}

\theoremstyle{plain}

\theoremstyle{definition}

\theoremstyle{remark}

\newcommand{\eg}{\textit{e.g.}}
\newcommand{\ie}{\textit{i.e.}}

\usepackage{kotex}

\icmltitlerunning{DNNs May Determine Major Properties of Their Outputs Early, with Timing Possibly Driven by Bias}

\begin{document}

\twocolumn[
\icmltitle{Deep Neural Networks May Determine Major Properties of \\ Their Outputs Early, with Timing Possibly Driven by Bias}

\icmlsetsymbol{equal}{*}

\begin{icmlauthorlist}
\icmlauthor{Song Park}{equal,na}
\icmlauthor{Sanghyuk Chun}{equal,na}
\icmlauthor{Byeongho Heo}{na}
\icmlauthor{Dongyoon Han}{na}
\end{icmlauthorlist}

\icmlaffiliation{na}{NAVER AI Lab}

\icmlcorrespondingauthor{Song Park}{song.park@navercorp.com}
\icmlcorrespondingauthor{Sanghyuk Chun}{sanghyuk.c@navercorp.com}

\icmlkeywords{Machine Learning, ICML}

\vskip 0.3in
]

\printAffiliationsAndNotice{\icmlEqualContribution} %

\begin{abstract}
This position paper argues that deep neural networks (DNNs) mostly determine their outputs during the early stages of inference, where biases inherent in the model play a crucial role in shaping this process. We draw a parallel between this phenomenon and human decision-making, which often relies on fast, intuitive heuristics. Using diffusion models (DMs) as a case study, we demonstrate that DNNs often make early-stage decision-making influenced by the type and extent of bias in their design and training. Our findings offer a new perspective on bias mitigation, efficient inference, and the interpretation of machine learning systems. By identifying the temporal dynamics of decision-making in DNNs, this paper aims to inspire further discussion and research within the machine learning community.
\end{abstract}

\section{Introduction}

How do artificial deep neural networks (DNNs) determine their outputs? What is the inner mechanism of the inference of DNNs? Despite the importance of this question, we still know very little about their inference mechanism.
This question becomes particularly intriguing when comparing DNNs to human decision-making systems. Do DNNs make decisions through deliberate, iterative reasoning, or do they arrive at their outputs almost instantaneously during inference? While these questions are challenging to answer, human decision-making offers some interesting analogies.

Machine learning (ML) researchers often assume that humans are rational and logical, while machines are biased and less reliable. However, extensive research from cognitive science and psychology supports that human decisions are not purely rational. Instead, humans often rely on intuition \citep{haidt2001emotional,kahneman2002maps,Kahneman2003-KAHAPO,gigerenzer2011heuristic} and emotion \citep{slovic2007affect,jarcho2011neural} as heuristics during the early stages of decision-making, with rationality serving to justify outcomes post-hoc \citep{Kahneman2003-KAHAPO,evans2008dual,gigerenzer2011heuristic}. Heuristics is fast and efficient but prone to errors, while rationality is slower and more deliberate.
This position paper suggests a hypothesis that this dual-process theory for human decision-making systems may coincide with the inner mechanism of DNN inference.

\begin{figure*}[ht!]
    \centering
    \includegraphics[width=.93\linewidth]{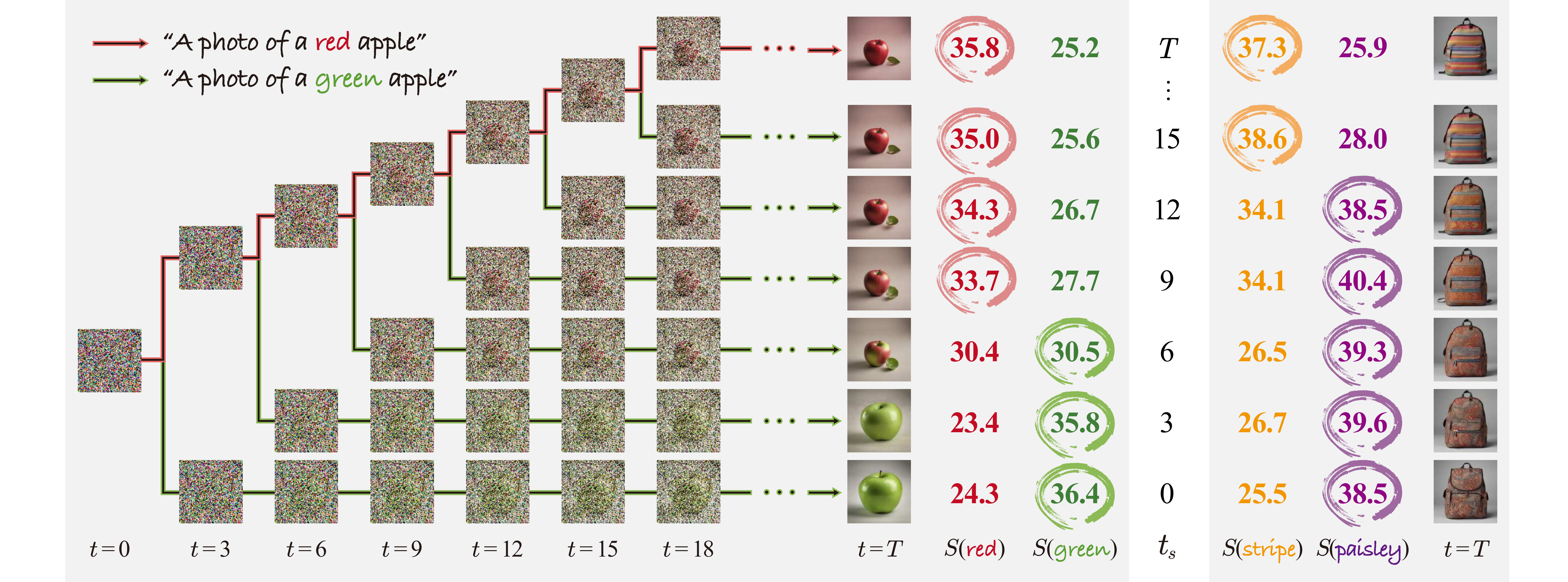}
    \vspace{-.5em}
    \caption{\small {\bf Overview of the proposed framework.} We choose two prompts (initial prompt $c_i$ and altered prompt $c_a$) formatting ``A photo of \texttt{[attribute]} \texttt{[entity]}'', where two prompts have the same \texttt{[entity]} but different \texttt{[attribute]}. At the timestamp $t_s$, we alter the initial text condition $c_i$ to the new condition $c_a$. Then, we measure the impact of each prompt using the CLIP similarity between the generated image and text prompts. We can observe that there exists a ``switching point'' where the generated image is influenced more to $c_a$ rather than $c_i$ (\eg, 9 for the apple example and 15 for the backpack example). Different attributes show different switching points, whereas a more biased attribute has an earlier one (\eg, the left color example shows an earlier conversion than the right pattern example).
    }
    \label{fig:teaser}
\end{figure*}

In this position paper, we hypothesize that \textbf{DNNs may determine their outputs during the early stage of the inference process, with the timing of this determination may depend on their ``heuristics'' (or in a more ML-related term, ``bias'' or ``shortcut'')}.
Specifically, we argue that DNNs rely on early-stage ``intuitive'' mechanisms, analogous to human heuristics, to quickly fix key aspects of their outputs. The remaining stages of inference serve to refine and finalize these initial decisions. Furthermore, we suppose that the timing of this early-stage determination is modulated by the model's bias toward specific features. For example, a model heavily biased toward color may fixate on color features earlier in the inference process than on other attributes, such as shape \citep{geirhos2018stylized_imagenet}.

To explore this hypothesis, we analyze the inference process of large-scale generative models (GMs) which have been attracting significant attention not only for their impressive generation quality but also for their potential connections to human intelligence. These models demonstrate emergent properties such as creativity, contextual understanding, and flexibility, which were traditionally considered unique and special properties of human cognition. By examining how powerful GMs estimate outputs, we can gain insights into both the strengths and limitations of machine intelligence.

Specifically, we study the inference mechanism of diffusion models (DMs), which generate outputs iteratively and provide a temporal trajectory of decision-making \citep{ho2020ddpm,song2021ddim}. By focusing solely on the inference process of pre-trained DMs, we eliminate confounding factors introduced during training, allowing us to isolate and analyze how these models ``determine'' their outputs at each step of the generation process. The step-by-step iterative mechanism of DMs makes them particularly suitable for studying the timing and dynamics of output determination, as they provide a temporal trajectory of decision-making rather than a single forward pass seen in conventional DNNs. Furthermore, their ability to understand high-level inputs like language prompts enables a more human-understandable framework for studying inference behavior.

We investigate how quickly text-to-image (T2I) DMs fix their decisions during the iterative process. As illustrated in \cref{fig:teaser}, we first guide the model with an initial prompt (\eg, ``a photo of a red apple'') and alter the prompt in the middle of the diffusion process (\eg, ``a photo of a green apple''). We measure whether the generated images follow the initial or altered prompt to determine the ``timing'' of the decision-making. Conceptually, if we alter the prompt at the first step, the generated image will be aligned to the altered prompt (\ie, ``green apple'' as shown in the $t_s=0$ example). On the other hand, if we alter the prompt at the later diffusion process, the generated image might not consider the altered prompt but simply follow the initial prompt (\ie, ``red apple'' as shown in the $t_s=15$ example). There might be a ``switching point'' where the generated image follows the altered prompt rather than the initial prompt; we define this switching timing as the moment of the ``decision-making'' with ``heuristic''. If this timing is closer to the early stages, it would suggest that the final output is already determined very early in the process. Conversely, if the change occurs closer to the later stages, it would indicate that the output is generated with deliberation.
Furthermore, as shown in \cref{fig:teaser}, we observe that the timing becomes later if we use a less biased attribute, \eg, the backpack pattern example shows later switching than the apple color example.

In our experiments, we examine five state-of-the-art T2I DMs and show that most models determine their outputs in the early inference stage (\eg, around 5 steps among 50 diffusion steps). Furthermore, if we use a more biased cue (\eg, color), a model tends to fix their output earlier than a less biased cue (\eg, material). For example, when we use color prompts, SD1.4 tends to ``switch'' the predicted output at step 7, while when we use material prompts, the timing becomes around 30. We observe that this tendency of the hasty determination and the bias-related timing happens regardless of the choice of the models.

\begin{figure*}[ht!]
    \centering
    \includegraphics[width=0.93\linewidth]{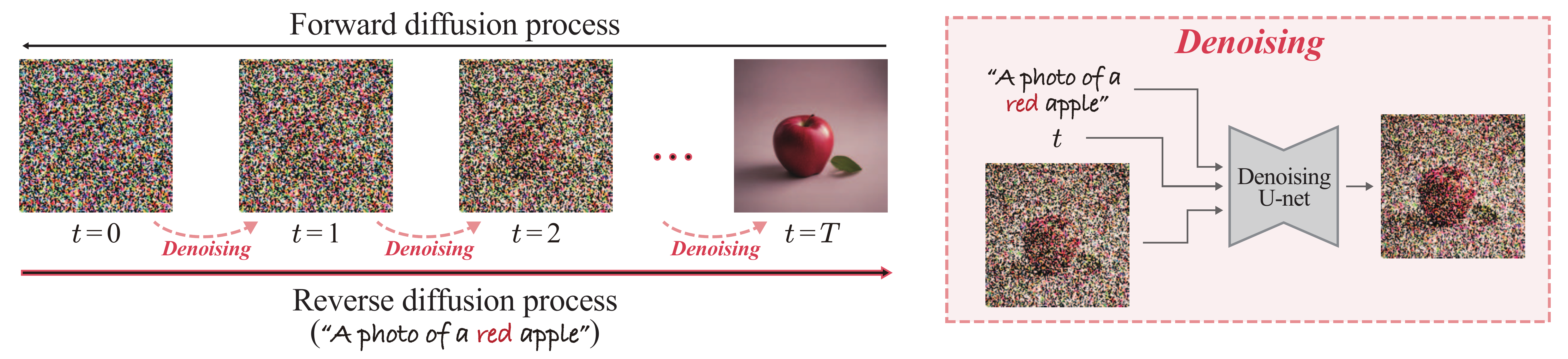}
    \vspace{-.75em}
    \caption{\small 
    The inference process of DMs (\ie, reverse process) is tractable over each intermediate output and each step is controllable by a flexible and human-understandable text prompt. We examine the temporal dynamics of inference using this iterative inference process.}
    \label{fig:diffusion_explanation}
\end{figure*}

\section{Related Work}

\subsection{Human decision-making system}

Human decision-making system is a complex interplay between intuitive and deliberative processes. \citet{haidt2001emotional} suggested that human judgments, particularly moral ones, are dominantly driven by intuitive processes, with reasoning often serving as a post hoc justification. \citet{kahneman2002maps} further elaborated on this dual-process theory, distinguishing between System 1 (fast, automatic, and intuitive) and System 2 (slow, effortful, and analytical) processes. System 1 dominates most everyday decision \citep{evans2008dual} and enables humans to make rapid decisions, often relying on efficient heuristics, but can introduce biases \citep{gigerenzer2011heuristic}. This heuristics can be intuition \citep{evans2008dual} or emotion \citep{slovic2007affect}.
These works highlight that human decision-making is not purely rational but deeply influenced by intuition, emotion, and heuristics.
In this paper, we suppose that artificial systems may also have a similar mechanism with humans.

\subsection{Machine decision-making system and their bias}

DNN inference mechanism has been widely studied, mostly focusing on their hierarchical behavior. \citet{zeiler2014visualizing} demonstrated that DNNs behave as sequential feature extractors, with earlier layers capturing low-level features such as edges and textures, and deeper layers focusing on more complex patterns and object parts. Similarly, the information bottleneck theory \citep{tishby2015deep,michael2018information_bottleneck} explains that earlier layers compress the input by removing redundant features, while later layers focus on prediction. While these approaches provide valuable insights for DNN inference, they assume unified and consistent behavior regardless of input properties.

Recent research highlights that DNNs are inherently biased or rely on ``shortcuts'' \citep{geirhos2020shortcut}, namely, DNNs prefers simpler features (\eg, color or texture) over more complex ones (\eg, shape) \citep{geirhos2018stylized_imagenet}.
Although an architectural difference can make a minor change \citep{brendel2019bagnet,bahng2019rebias,naseer2021intriguing}, as shown by \citet{scimeca2022shortcut}, these biases exist regardless of the network architecture. Furthermore, certain cues (\eg, color) are preferred to other more complex ones (\eg, shape), highlighting that DNNs are inherently more likely to be biased toward features that are computationally simpler.
This paper supposes that this preference behaves similarly to ``fast heuristics'' in DNNs, enabling efficient but potentially error-prone decision-making during early inference stages.

\section{Preliminary: Diffusion Models}

Diffusion models (DMs) \citep{ho2020ddpm,song2021ddim} are a class of generative models (GMs) that iteratively refine noise to predict outputs. In the forward process of DM, noise is iteratively added to input over multiple steps (\cref{fig:diffusion_explanation} ``forward process''). The reverse process incrementally denoises the corrupted data by a network, reconstructing the original input (\cref{fig:diffusion_explanation} ``reverse process'').
More specifically, we generate an output by the reverse process, $p_\theta(x_T) := p(x_0)\prod_{t=1}^{T} p_\theta(x_t | x_{t-1})$, where $x_0$ denotes a random Gaussian noise, the first step of inference, and $x_T$ denotes an image, the last step.\footnote{Note that it is a convention to use $t=0$ for the original image and $t=T$ for the noise space. However, this paper uses $t$ as the step of inference, \ie, $t=0$ for random noise (the first step of the inference) and $t=T$ for image (the last step of the inference).}
Namely, from a random Gaussian noise, a DM iteratively predicts the next output $T$ times to estimate the distribution of data $x$. We suppose that each diffusion step denotes the ``stage'' of final decision-making, where the total number of stages is $T$ (\ie, the number of diffusion steps).
Unless specified, we set the diffusion step to 50 for all experiments.

A key advantage of DMs lies in their iterative generation process, which allows for explicit control over the generation steps. This iterative nature is beneficial for analyzing the decision-making dynamics, as each step provides a snapshot of the intermediate stages of the model's output. Specifically, a text-conditioned DM uses text embeddings from the pre-trained models (\eg, CLIP \citep{radford2021clip}) for the reverse process, producing outputs aligned with the given text prompt (\cref{fig:diffusion_explanation}b). By using a natural language condition, we can use a human-understandable condition to control each intermediate stage of the model's output.
These properties make DM easier to analyze by conflicting a prompt and observing how the model output aligns with human-understandable cues. We use this iterative inference process as a proxy of the temporal dynamics of machine inference, resembling human reflection or deliberation processes. In the following experiments, we will inspect whether DNN inference is mostly dominated by early-stage or distributed more evenly across the iterative process.

\section{DNNs Determine Their Outputs in the Early Stages of Inference, Influenced by Bias}

\subsection{Experiment design}

The overview of our experiment is illustrated in \cref{fig:teaser}.
Assume we have two different text prompts, an initial prompt $c_i$ (\eg, ``red apple'') and an altered prompt $c_a$ (\eg, ``green apple''). We start generation with $c_i$ until a timestamp $t_s \in [0, T]$, where $x_0$ equals a random Gaussian noise. From $t_s$, we change $c_i$ to $c_a$ and generate an image $x^{t_s}$, \ie, $x^T$ equals to an image solely guided by $c_i$ and $x^0$ equals to an image guided by $c_a$.
We then analyze how the final generated image $x^{t_s}$ reflects $c_i$ and $c_a$ (\eg, check whether the generated apple is red or green). More specifically, we quantify the impact of each prompt using the CLIP image-text similarity function $S(x^{t_s}, c_i)$ and $S(x^{t_s}, c_a)$. If $S(x^{t_s}, c_i)$ is larger than $S(x^{t_s}, c_a)$, we may assume that the network already determines the final output with $c_i$ at timestamp $t_s$.

\begin{figure}[h]
    \centering
    \includegraphics[width=0.9\linewidth]{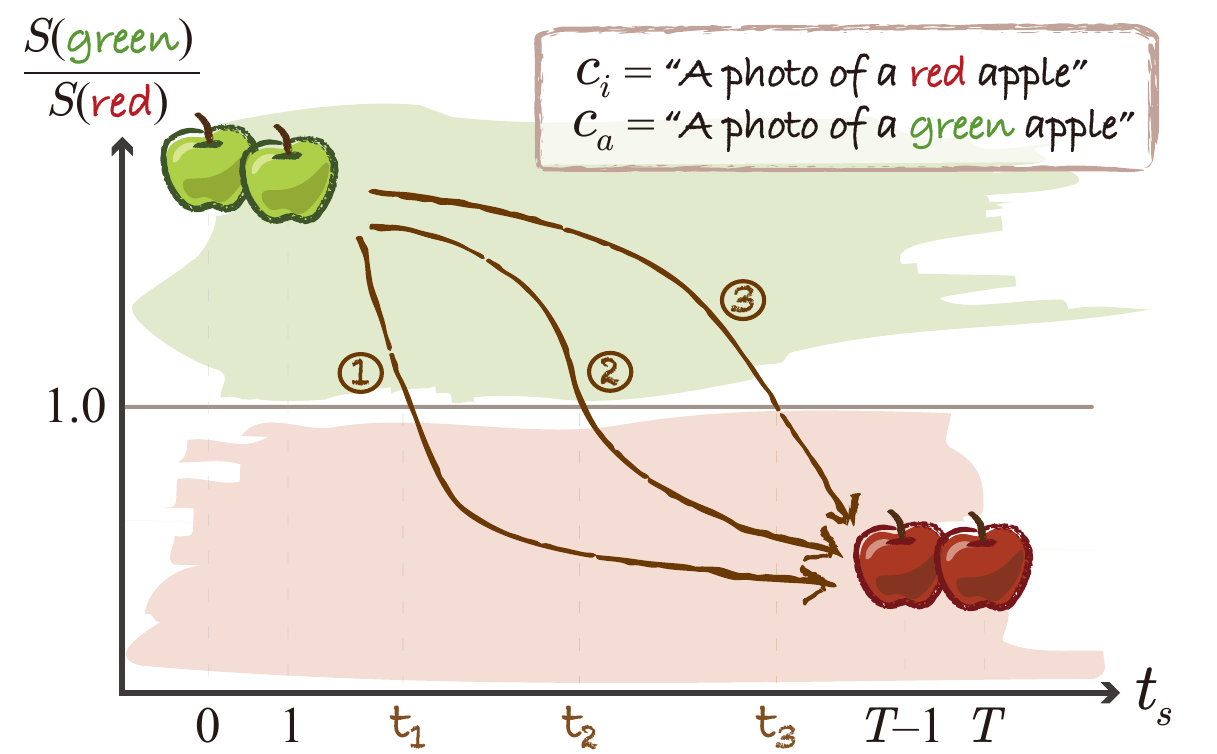}
    \caption{\small We show examples of $x^{t_s}$ by varying $t_s$ from $0$ to $T$ and their estimated CLIP scores. The x-axis denotes $t_s$, the timestamp where the initial prompt $c_i$ is changed to the altered prompt $c_a$. The y-axis denotes the ratio of $S(x^{t_s}, c_i)$ and $S(x^{t_s}, c_a)$; higher means the generated image is more influenced by $c_a$ and vice versa. When will the generated image be more influenced by $c_a$ than $c_i$? If the output is more influenced by $c_i$, then we need a smaller $t_s$ to make the image more influenced by $c_i$ (\eg, the $t_1$ case). Otherwise, a larger $t_s$ will be sufficient (\eg, the $t_3$ case).}
    \label{fig:switching_point_example}
\end{figure}

We define the ``switching point'' $t_s^\prime$ is the smallest timestamp where $S(x^{t_s^\prime}, c_i) > S(x^{t_s^\prime}, c_a)$. If the switching point $t_s^\prime$ is closer to the start of the inference process, it suggests that the model determines the major properties of the generated image at an early stage. Conversely, if $t_s^\prime$ occurs closer to $T$, it implies that more inference steps are required to finalize the output.
For example, the $t_1$ case of \cref{fig:switching_point_example} determines the output earlier than $t_2$ and $t_3$ cases.
By measuring $t_s^\prime$ under various conditions, we support our hypothesis that \textbf{DNNs may determine their outputs during the early stage of the inference process, with the timing of this determination being influenced by their inherent biases.}

We evaluate five text-conditioned DMs: Stable Diffusion 1.4 \citep{rombach2022latent_diffusion}, Stable Diffusion XL \citep{podell2023sdxl}, Stable Diffusion 3 \citep{esser2024sd3}, Kandinsky 3 \citep{arkhipkin2023kandinsky}, and Karlo UnCLIP \citep{kakaobrain2022karlo-v1-alpha}, considering their architectural differences. We use the pre-trained weights available from HuggingFace.
We describe more details in \cref{appendix:sec:diffusion_model_details}. For the CLIP similarity, we use ViT-H-14 CLIP trained by \citet{fang2023dfn}.

We consider text prompts in the format ``A photo of a \texttt{[attribute]} \texttt{[entity]}'' in two distinct scenarios. In the first scenario, \texttt{[attribute]} corresponds to \textit{color}, \textit{pattern}, \textit{shape}, and \textit{material}, while \texttt{[entity]} represents one of \textit{10 common objects} that shows minimal bias for the given attributes. In the second scenario, \texttt{[attribute]} refers to \textit{gender} and \textit{ethnicity}, and \texttt{[entity]} corresponds to \textit{16 professions} chosen to include diverse contexts and demographic representations.
For both scenarios, we measure the switch timing between prompts with the same entity but different attributes, \eg, ``red apple'' and ``green apple''.

\paragraph{Scenario 1. Common objects.}
We use four visual attribute groups: color (10 attributes, \eg, ``red'' or ``green''), pattern (7 attributes, \eg, ``stripes'' or ``paisley''), shape (6 attributes, \eg, ``round'', ``square''), and material (8 attributes, \eg, ``fabric'' or ``metal''). For each attribute group, we choose 10 objects that is minimally biased to the attribute group (\eg, ``pen'' for color, ``backpack'' for pattern, and ``bowl'' for material) -- the full list is in \cref{appendix:sec:full_list_attribute_entity}.
For each attribute type, we randomly select ten pairs of attributes (\eg, red and green) and generate five different sets for each object, where each set contains generated image $x^{t_s}$ from $t_s=0$ to $t_s=T=50$. Namely, we generate $10\times5\times50\times10=25,000$ images for each attribute. In our experiment, we have four attribute groups and five models, hence, 500k generated images are used for analysis.

\paragraph{Scenario 2. Humans.}
Following StableBias \citep{luccioni2024stable_bias}, we choose gender (male, female, and non-binary) and ethnicity (black, white, asian, and hispanic) as the altering attributes, \ie, \texttt{[attribute]}.\footnote{We acknowledge that these attributes cannot represent all human beings and some attributes can be even inadequate. However, we clarify that we chose the terms from StableBias \citep{luccioni2024stable_bias} with a careful initial study. We will clarify more details in Impact Statement section.} We also choose 16 professions as \texttt{[entity]}, which show the most and the least diverse generation results across genders and ethnicities from StableBias. For example, \citet{luccioni2024stable_bias} showed that DMs generate the most diverse images for ``singer'' and the least diverse ones for ``tractor operator''. The full list of 16 professions can be found in \cref{appendix:sec:full_list_attribute_entity}.
Similar to scenario 1, we generate ten set of images for each pair of attributes (gender has six valid pairs and ethnicity has 12 valid pairs) and each profession. Namely, we generate $(6+12)\times10\times50\times16=144,000$ images for each model, and overall 720k generated images are used for analysis.

\begin{figure*}[ht!]
    \centering
    \includegraphics[width=\linewidth]{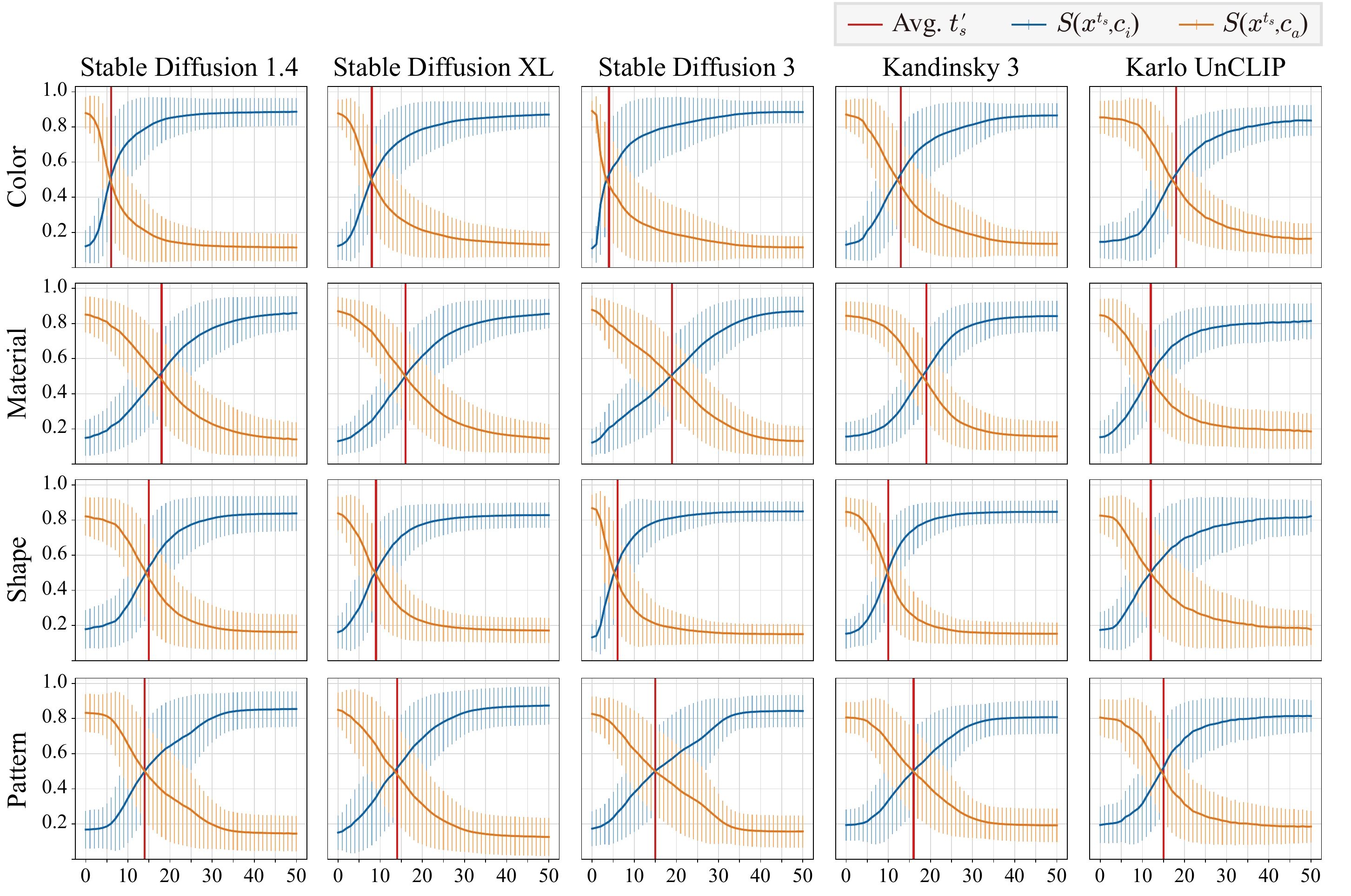}
    \vspace{-2em}
    \caption{\small 
    {\bf DNNs may determine the major properties of their output at an early stage.} We plot the average and the standard error of \textcolor{blue}{$S(x^{t_s}, c_i)$} and \textcolor{orange}{$S(x^{t_s}, c_a)$}. $S(x^{t_s}, c)$ denotes a CLIP similarity between a text prompt $c$ and a generated image $x^t_s$ by altering the initial prompt $c_i$ to altered prompt $c_a$ at timestamp $t_s$.
    $x^{0}$ equals to an image fully conditioned by $c_a$ and $x^{50}$ equals to one conditioned by $c_i$ (\cref{fig:switching_point_example} shows an example).
    Each point is computed with 50 samples (10 attribute pairs and 5 random seeds). 
    The red line is the ``switching point'', the smallest $t_s'$ where $S(x^{t_s'}, c_a) > S(x^{t_s'}, c_i)$ on average, which is a proxy of the timing of the ``determination''.
    }
    \label{fig:exp_main_switch}
    \vspace{-.5em}
\end{figure*}

\begin{figure*}[ht!]
    \centering
    \vspace{-0.5em}
    \includegraphics[width=\linewidth]{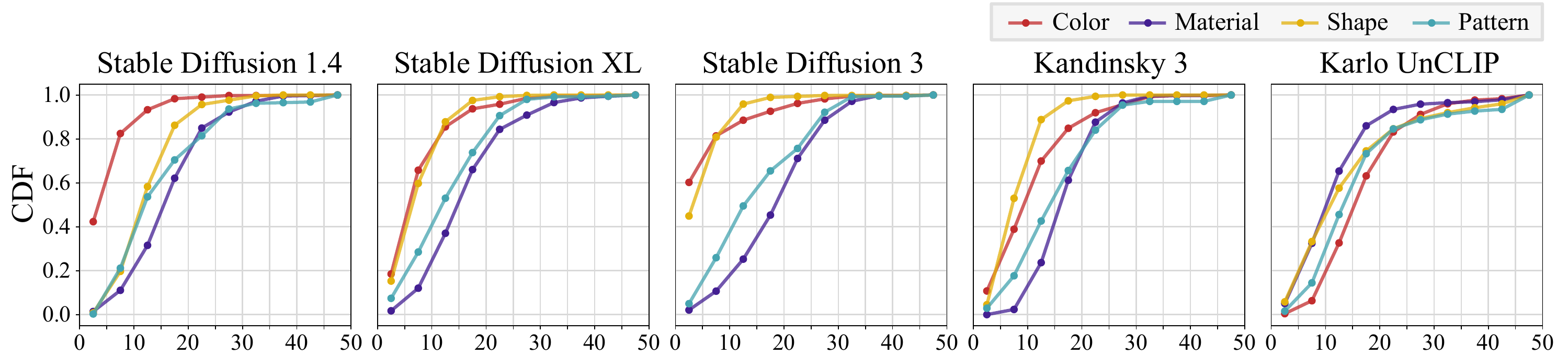}
    \vspace{-2em}
    \caption{\small {\bf Cumulative histogram of the sample-wise switching timing for each model and attribute.} 
    Note that we have ten objects, ten attribute pairs, and five random seeds; hence, each histogram contains 500 samples.}
    \label{fig:exp_main_switch_points}
    \vspace{-1em}
\end{figure*}

\begin{figure*}[ht!]
    \centering
    \includegraphics[width=\linewidth]{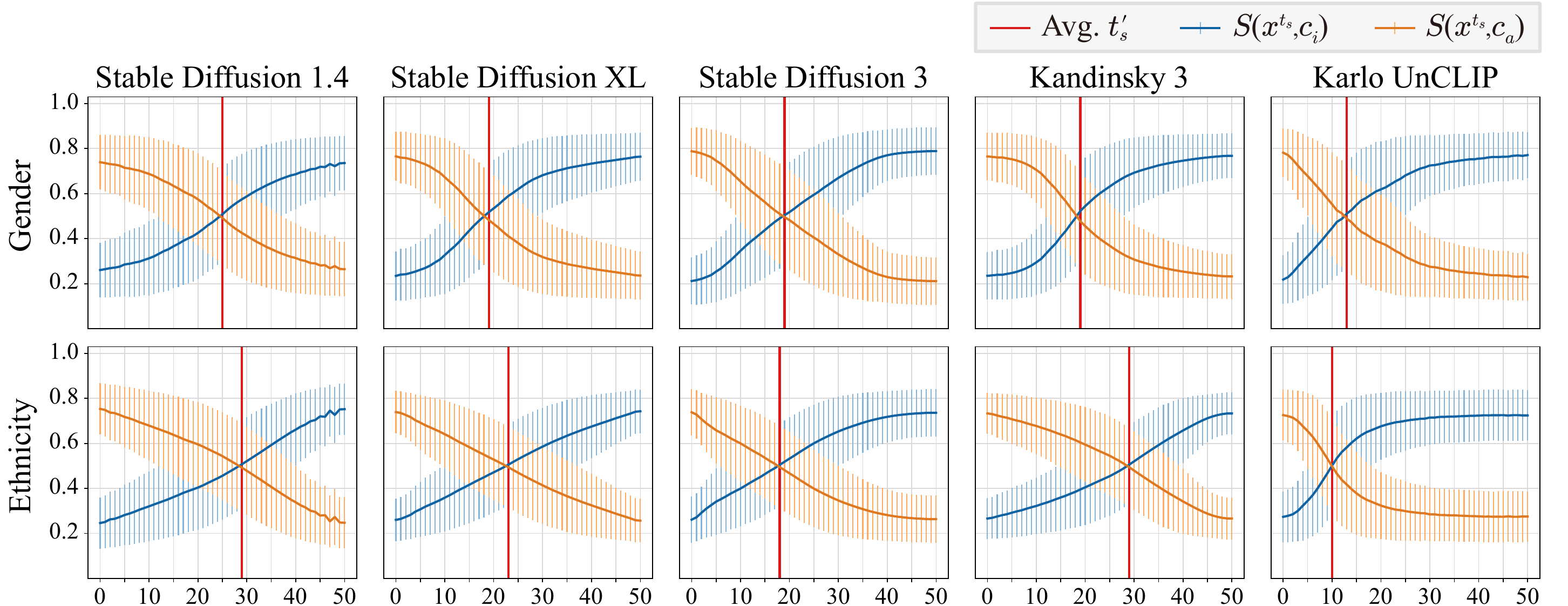}
    \vspace{-2em}
    \caption{{\bf Switching point for human attributes.} The details are the same as \cref{fig:exp_main_switch}.}
    \label{fig:exp_main_human}
\end{figure*}

\begin{figure*}[ht!]
    \centering
    \includegraphics[width=\linewidth]{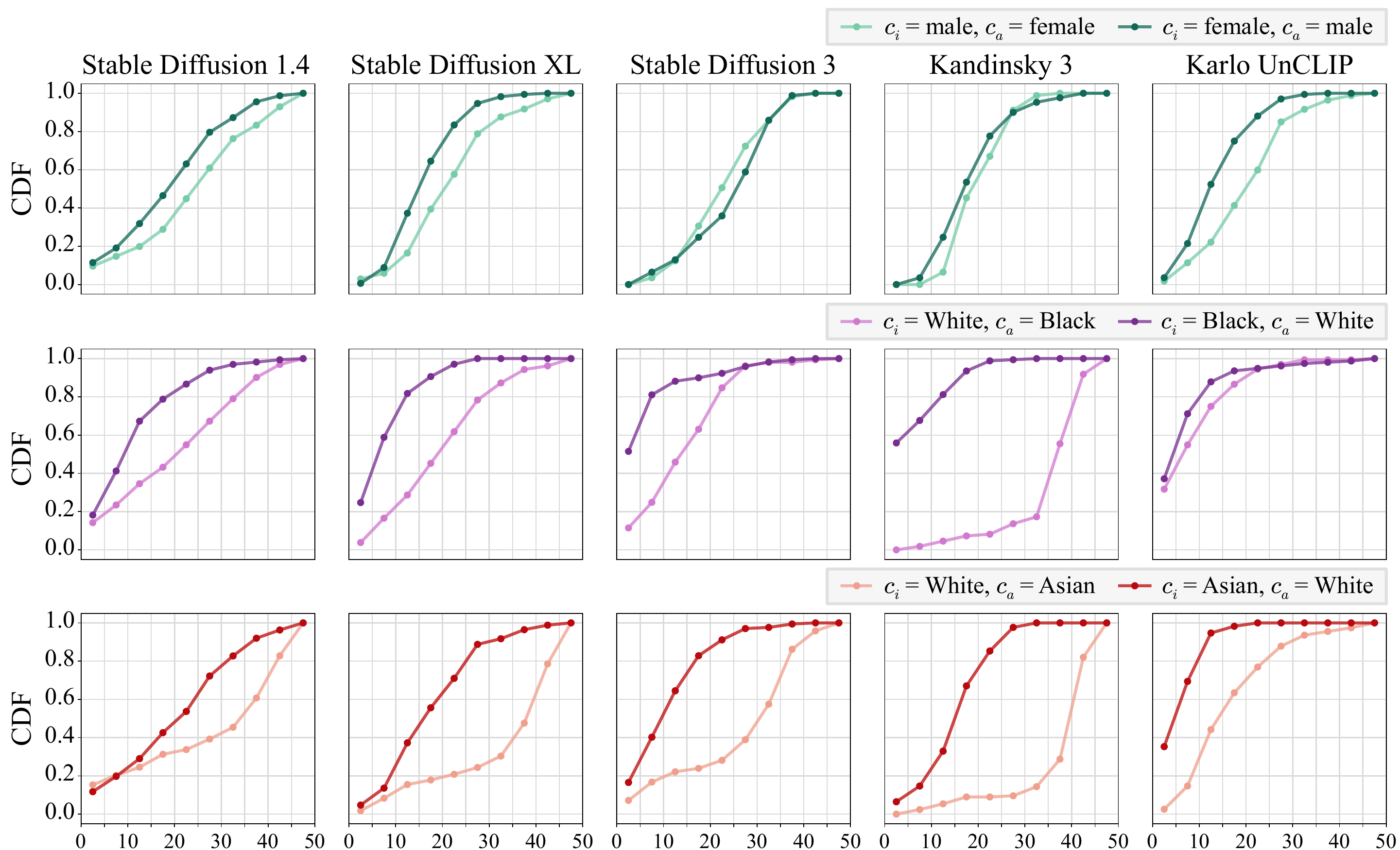}
    \vspace{-2em}
    \caption{\small {\bf Cumulative histogram of the sample-wise switching timing for human attributes.} 
    The details are the same as \cref{fig:exp_main_switch_points}.}
    \label{fig:exp_human_switch_points}
\end{figure*}

\subsection{DNN outputs are determined at an early stage}

\cref{fig:exp_main_switch} shows the average of $S(x^{t_s}, c_i)$ as \textcolor{blue}{blue lines} and the average of $S(x^{t_s}, c_a)$ as \textcolor{orange}{orange lines} from $t_s=0$ to $50$ for five models and four attribute types (color, pattern, shape, and material) with standard errors.
\textbf{In most cases, the ``switching point'' $t_s'$ occurs very early in inference}, often within the first 15 steps, or even within the first 5 steps out of 50.
We also plot the histogram of $t_s^\prime$ for each model and attribute in \cref{fig:exp_main_switch_points}.
From the figures, we observe that for some settings, only very few steps are required to determine the property of the generated outputs. For example, Stable Diffusion 3 fixes its output for 60\% of generated images with color attributes within just five steps (\cref{fig:exp_main_switch_points}).
\textbf{However, even for the same model, more steps are required to determine the outputs with a more ``difficult'' attribute.} For example, Stable Diffusion 3 with material attributes requires over 20 steps to reach the same threshold.

While most of the models show the gap between switch timing measured by easy features (\eg, color) and difficult features (\eg, material), we observe that the Karlo UnCLIP model shows a smaller gap compared to the others. We presume that this is because UnCLIP has two modules taking separate text conditions; the prior model and the decoder model. We only control the text condition on the prior model, while the decoder model only takes the initial prompt $c_i$. We plot the case when the decoder model is controlled while the prior model only uses $c_i$ in \cref{appendix:sec:more_experiment_unclip_decoder}

\cref{fig:exp_main_human} and \cref{fig:exp_human_switch_points} show that the models behave for human attributes similarly to the results of common objects.
\cref{fig:exp_main_human} shows that the switching happens at an early stage as common object examples. We found that their average switching timing would not be as early as common objects, but if we focus on specific attributes, we can still observe similar phenomena.
Specifically, \cref{fig:exp_human_switch_points} shows that \textbf{the switching timing is also affected by how the model is initially biased to a specific attribute.} For example, the first row shows that most models are male-biased, \ie, male images are easier to generate, but female images are more easily altered by the male prompt. More significantly, this gap becomes even larger for ethnicity attributes; the models are severely biased toward a specific ethnicity.

\subsection{The timing of early determination may be dominated by inherent bias of the model}

\begin{figure}[t]
    \centering
    \begin{subfigure}[b]{0.47\linewidth}
        \centering
        \includegraphics[width=\linewidth]{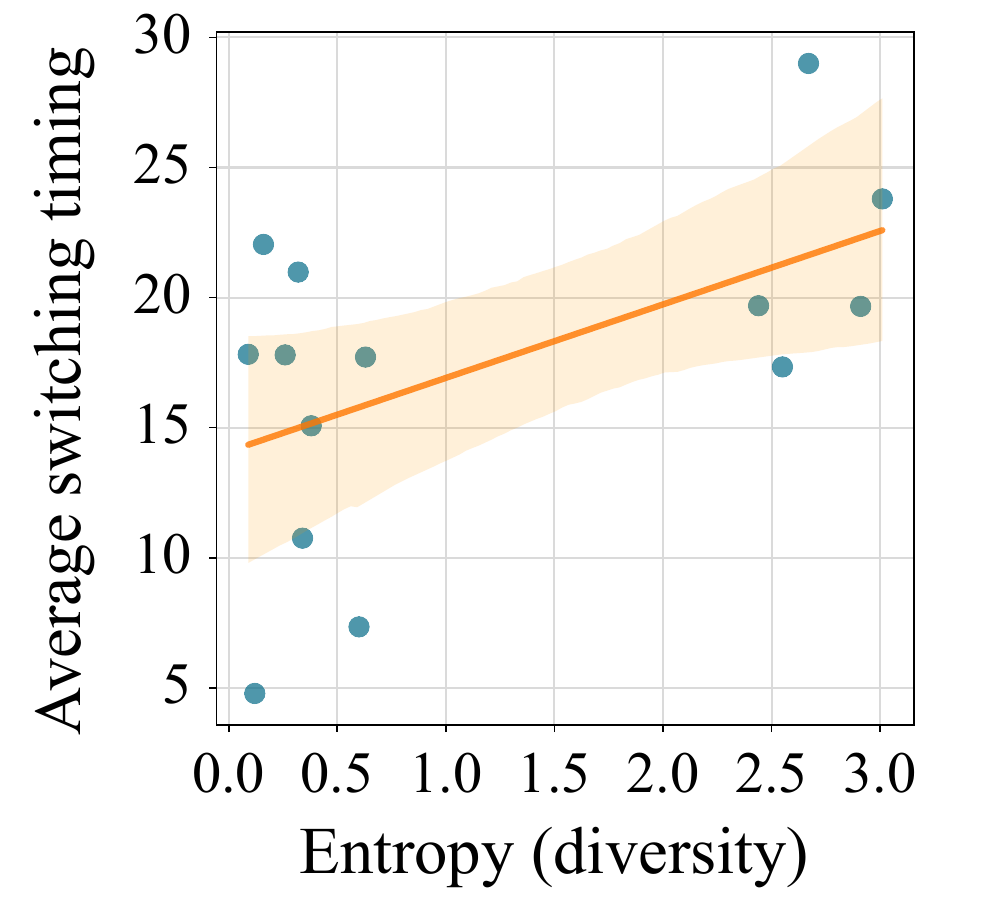}
        \caption{\small {\bf Entropy vs. switching timing for human attributes.}}
        \label{fig:entropy_vs_switching_point_human_sd1_4}
    \end{subfigure}\hfill
    \begin{subfigure}[b]{0.47\linewidth}
        \centering
        \includegraphics[width=\linewidth]{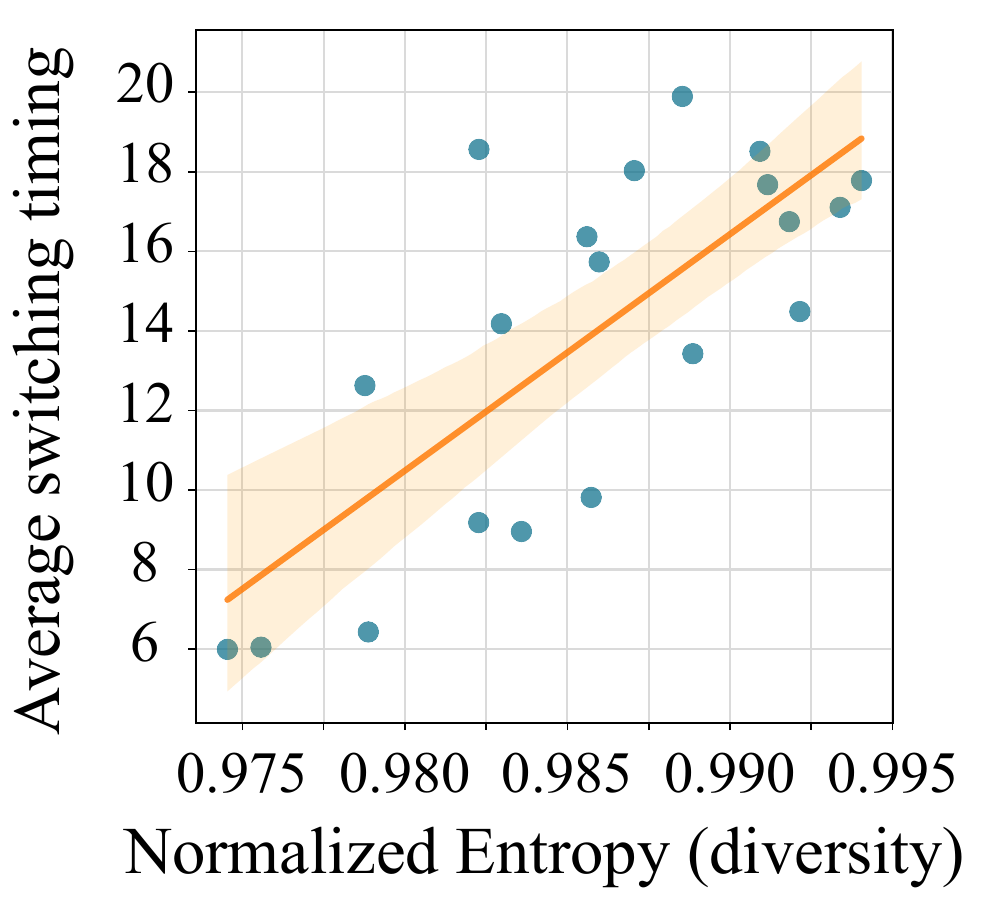}
        \caption{\small {\bf Normalized entropy vs. switching timing for objects.}}
        \label{fig:entropy_vs_switching_point_objects}
    \end{subfigure}
    \caption{\small {\bf Diversity vs. determination timing.} We plot the relationship between the diversity measure of the generated images and the early determination timing for (a) human attributes and (b) common objects. We use normalized entropy for common objects due to the numbers of attributes are different by their types.}
    \label{fig:entropy_vs_swithing_point_main}
\end{figure}

Why does early determination occur? Why is there a gap between the determination timings for different attributes?
In this subsection, \textbf{we hypothesize this is because of the inherent bias of the models} and empirically support this claim. Namely, if a model shows a more biased behavior for a specific attribute (\eg, color), its switching timing will become earlier than a non-biased attribute (\eg, material).

We first verify this hypothesis with the bias in human generation found by StableBias \citep{luccioni2024stable_bias}. StableBias provides a diversity measure of generated images for specific professions and models, where the diversity is measured by the prediction entropy on pre-defined clustering. We plot the relationship between gender generation diversity and the average switch timing for each profession in \cref{fig:entropy_vs_switching_point_human_sd1_4}. Interestingly, there exists a positive correlation between the generation diversity and the average switching timing, \textit{which supports our hypothesis}. Note that among the models used in our analysis, only SD1.4 results are provided from StableBias.
We additionally verify our claim with common object images for more diverse models, \ie, five models used in our experiments.

We generate 100 images for each object we used for the experiments (The list can be found in \cref{appendix:tab:scenario_one_object_list}) with prompts ``a photo of a \texttt{[entity]}''; hence, each attribute has 1,000 images.
Then, we measure the CLIP similarity between the generated images and the attributed prompts (\ie, ``a photo of a \texttt{[attribute]} \texttt{[entity]}''). Using this similarity score, we compute the zero-shot prediction entropy of the generated images to measure the generation diversity. We use normalized entropy (\ie, $\frac{-p\log p}{\log d}$, where $d$ is the dimension of $p$) to minimize the impact of the attribute numbers (\eg, we have 10 colors and 6 shapes. This will change the scale of their entropy).

In \cref{fig:entropy_vs_switching_point_objects}, we report the relationship between the generation diversity (measured by the normalized entropy) and the early-determination timing (measured by the average switching timing in \cref{fig:exp_main_switch}) for all model-attribute pairs. Interestingly, we found a positive correlation between the diversity and the early-determination timing. In other words, if a model shows a more biased behavior to a specific attribute, then the model will determine the main property of the generated image conditioned by the biased attribute. \textit{This again empirically supports our hypothesis.}

\paragraph{Conclusion.}
In this section, we empirically show that DNNs determine the major properties of their outputs at a very early moment of inference (\eg, less than 5 for specific cues) with two scenarios (common objects as shown in \cref{fig:exp_main_switch} and human attributes as shown in \cref{fig:exp_main_human}). Furthermore, we show that this timing of the determination is highly correlated to how the model is biased toward the given cue in \cref{fig:entropy_vs_swithing_point_main} (\eg, when we generate images with a more biased cue, such as color, the model determines the output much earlier than a less biased cue, such as material).

\section{Alternative Views}
\paragraph{How can our claim be extended to non-iterative or non-generative models?}
One opposing perspective can arise from the specificity of our findings to diffusion models (DMs). While DMs are an ideal case for studying iterative inference processes, one may argue that the observations from DMs would not generalize to other architectures. For example, feedforward DNNs operate in a single forward pass, lacking the iterative nature that DMs leverage. As a result, the insights about early-stage determination or the effects of determination timing may not translate to architectures that are fundamentally different in their inference mechanisms.
We recognize that our findings are grounded in DMs, but we view these as a case study to explore broader patterns that could inform future research across methods.

Another class of models worth considering is auto-regressive (AR) models, such as language models (LMs). Although their inference mechanism consists of multiple forward passes, modifying conditions mid-inference (as we did in DMs) is not straightforward in LM inference. Additionally, evaluating generated texts in a controlled and measurable way is inherently challenging. Due to these factors, we did not include AR models in our primary analysis. While investigating early-stage determination in AR models would be valuable, it falls outside the scope of this work.

\paragraph{Causality-aware models and concept bottleneck models may show different behaviors.}
Another critique can arise from the fact that certain models, such as concept bottleneck models \citep{koh2020concept} or causal models \citep{kaddour2022causal}, inherently rely on explicit intermediate concepts or causal mechanisms to make decisions. These models are designed to ensure that decision-making is interpretable and structured in a way that is consistent across inputs. Unlike the dynamic mechanisms described in this paper, these models would not rely on early-stage ``heuristics''. Therefore, our analysis may not lead to the same result for these models.
However, these models are not universally adopted across domains and most state-of-the-art models, hence, this argument can be limited to a specific case.
Furthermore, we would assume that the mechanism for estimating the intermediate concepts or causal nodes behaves similarly to general DNNs, which would follow our hypothesis.

\paragraph{Bias would not work as heuristics.}
Finally, some critics may argue that comparing DNN bias to ``fast heuristics'' of human decision-making systems oversimplifies the nature of machine learning models. While heuristic bias may be a useful analogy, DNNs may operate on statistical patterns in data, which can lead to biases that are not analogous to human intuition.
Specifically, many studies have argued that a biased behavior by DNNs originated from a biased dataset rather than their inherent property \citep{geirhos2020shortcut}. However, at the same time, some studies have suggested that the bias can be easily happened due to the simplicity bias \citep{scimeca2022shortcut}. Namely, even now, the origin of machine bias and its mechanism is known very little despite their importance.
In this paper, we do not directly suggest the mechanism beyond the inference, but we try to reveal the hidden behavior of DNNs; their outputs are determined during a very early inference stage, and the timing of the determination is correlated to how the model is biased. Identifying the actual mechanism will be an interesting future direction. We will discuss this in \cref{sec:discussion}.

\section{Discussion}
\label{sec:discussion}

In this position paper, we claim that \textbf{DNNs may determine their outputs at the early stage of the inference process}. Additionally, we argue that \textbf{the timing of this early determination may be influenced by biases inherent in the model}. In this section, we further explore the implications of these claims and discuss how they provide new insights and opportunities for improving DNNs.

\paragraph{Understanding the inner mechanism of DNN inference.}
While the existing studies focus on the hierarchical feature extraction process \citep{zeiler2014visualizing}, our study introduces a complementary perspective by focusing on the ``determination stage'' during inference. Unlike prior approaches, we hypothesize that DNNs may behave differently depending on the nature of their inputs, particularly the complexity of the features. This suggests that the inference process is not universally consistent but is dynamically modulated by the input.
One possible interesting future research direction could be a deeper understanding of the bias-related mechanism. For example, as discussed by \citet{scimeca2022shortcut}, this can be related to simplicity bias and the loss surface when we introduce biased features.

\paragraph{A new lens for bias mitigation and ``chain-of-thoughts''.}
If, as we suppose, bias behaves as ``fast heuristics'' similar to humans, we may devise a method to mitigate bias inspired by human cognitive processes. For example, \citet{haidt2001emotional} observed that humans, when afforded the opportunity for deliberation, shift from heuristic-driven decisions to more rational and accurate reasoning. This insight aligns with the concept of chain-of-thoughts (CoT) \citep{wei2022cot}, which enables large language models (LLMs) to engage in complex reasoning by following incremental, step-by-step logical prompts. Extending this analogy to DNNs, we would propose encouraging models to adopt intermediate reasoning steps during inference to reduce their reliance on shortcuts or biased features. For instance, introducing mechanisms that enforce iterative processing within generative models, such as multi-step deliberations in diffusion processes, could promote deeper and more balanced reasoning. This approach not only offers a framework for bias mitigation but also provides a new direction to better align machine reasoning with human-like reflective processes, improving both fairness and robustness in model outputs.

As a primitive study, we generate images with complex prompts with multiple features. In many cases, a DM cannot cover a complex prompt but only generates an image with selective features, mostly biased cues. In \cref{appendix:subsec:progressive_prompt}, we show qualitative examples when we control the model in a progressive manner, \ie, we start from the simplest one (\eg, ``a photo of a pajama'') then we update the prompt (\eg, ``a photo of a checkered pajama''). We expect that this direction can be helpful when a DM ignores a specific cue.

\paragraph{A new perspective on bias in inference mechanisms.}
Bias in DNNs is often viewed as a flaw, but we may argue that \textit{``DNN bias is not a bug, but a feature''}.
Similar to how humans rely on heuristics to make quick decisions in familiar situations, DNN biases may enhance efficiency when the given input is highly correlated with biased features.
Furthermore, if different architectures show different biased behaviors \citep{naseer2021intriguing}, leveraging diverse models could lead to improved performance as shown by \citet{hwang2024similarity}. This idea parallels findings in human decision-making, where groups with diverse individuals tend to outperform even the best individual within the group \citep{laughlin2006groups}.

We expect that this new perspective on the role of bias in inference can open a new direction for designing efficient and strong inference mechanisms based on input property.

\clearpage
\section*{Impact Statements}
This work investigates the role of bias in deep neural networks (DNNs) and explores how certain biases may function as efficient shortcuts in solving tasks. However, we strongly caution against prematurely concluding that bias is inherently beneficial, as such claims risk justifying discrimination within machine learning (ML) systems. Our analysis does not seek to justify biased decision-making but instead draws an analogy between DNNs and human cognitive processes, wherein heuristics serve as natural yet sometimes flawed mechanisms for problem-solving \citep{Kahneman2003-KAHAPO}.
In any case, our findings should not be used to justify biased or discriminatory behaviors in ML models. 

In our human attribute experiments, we use gender (male, female, and non-binary) and ethnicity (Black, White, Asian, and Hispanic) as attributes from StableBias \citep{luccioni2024stable_bias}. We acknowledge that these categories are limited and do not encompass the full diversity of human identities. Additionally, certain attribute definitions may themselves be inadequate or problematic. It is crucial to recognize that our study may introduce biases, and any application or extension of our results must carefully consider these limitations.

Furthermore, our findings suggest that some tasks may require deeper reasoning (\eg, more inference steps). However, this does not imply that simply increasing computational depth, such as slowing forward passes or using more parameters, leads to fairer or less biased outcomes. Specifically, our study cannot be used to justify that a more computationally expensive model is inherently less discriminatory. Bias in ML systems must be examined holistically, considering both algorithmic properties and the broader sociotechnical context.
Overall, we encourage the ML community to critically engage with these findings and to approach bias-aware modeling with careful ethical considerations.

\bibliography{main}
\bibliographystyle{icml2025}

\newpage
\appendix
\numberwithin{equation}{section}
\numberwithin{figure}{section}
\numberwithin{table}{section}
\onecolumn
\section*{Appendix}

\section{Experiment Design Details}
\label{appendix:sec:experiment_design}

\subsection{Details of prompt altering for diffusion models with multiple modules}
\label{appendix:sec:diffusion_model_details}

The Karlo UnCLIP model \citep{kakaobrain2022karlo-v1-alpha} has two separated modules, the prior module, and the decoder module, following Dall-E 2 \citep{ramesh2022hierarchical}. The prior module generates an image latent vector from the given text latent vector (both are extracted from CLIP \citep{radford2021clip}). After generating the image latent, the decoder module generates a pixel-level image. Here, both prior and decoder modules of Karlo UnCLIP are diffusion models and take a text condition for each step.
We empirically observe that the decoder text condition also affects a lot to the generated image (\ie, the decoder does not solely behave as ``decoder'', but it also behaves as a generative model). However, to make our analysis consistent, we let the decoder use the same text prompt $c_i$ while the prior module is controlled by our setting.

\subsection{Full List of Attributes and Entities}
\label{appendix:sec:full_list_attribute_entity}

\begin{table}[h]
    \small
    \centering
    \begin{tabular}{l|l}
    \toprule
    Attribute type & Attributes \\ \midrule
    Color & black, blue, brown, gray, green, pink, purple, red, white, yellow \\
    Pattern & argyle, camouflage, checkered, herringbone, paisley, polka dots, stripes \\
    Material & fabric, glass, leather, marble, metal, plastic, stone, wood \\
    Shape & round, square, triangular, hexagonal, star-shaped, heart-shaped \\
    \bottomrule
    \end{tabular}
    \caption{\small {\bf Attribute details for scenario 1.} For color attributes, we avoid the following highly similar pairs: (black, gray), (blue, green), (blue, purple), (brown, red), (brown, yellow), (green, yellow), (pink, purple), (pink, red).}
    \label{appendix:tab:scenario_one_attribute_list}
\end{table}

\begin{table}[h]
    \small
    \centering
    \begin{tabular}{l|l}
    \toprule
    Attribute type & Entities \\ \midrule
    Color & car, crayon, bicycle, notebook, pen, balloon, t-shirt, shoe, cup, button \\
    Material & table, chair, bottle, vase, cup, bowl, box, shelf, bench, tray \\
    Pattern & shirt, dress, jacket, scarf, sweater, skirt, pants, tote bag, backpack, wallet \\
    Shape & cookie, mirror, clock, pendant, coaster, birthday card, vase, badge, ornament, cake \\
    \bottomrule
    \end{tabular}
    \caption{\small {\bf Entities for each attribute for scenario 1.} For each attribute type, we choose ten objects minimally biased to the attribute.}
    \label{appendix:tab:scenario_one_object_list}
\end{table}

\begin{table}[h]
    \small
    \centering
    \begin{tabular}{l|l}
    \toprule
    Entity                     & Entropy from StableBias \citep{luccioni2024stable_bias} \\ \midrule
    singer                     & 2.85 \\
    cleaner                    & 2.51 \\
    aide                       & 2.30 \\
    teacher                    & 2.26 \\
    fast food worker           & 2.54 \\
    therapist                  & 2.03 \\
    hairdresser                & 1.92 \\
    air conditioning installer & 0.24 \\
    machinery mechanic         & 0.19 \\
    roofer                     & 0.15 \\
    drywall installer          & 0.14 \\
    tractor operator           & 0.09 \\
    dental assistant           & -    \\
    nutritionist               & -    \\
    receptionist               & -    \\
    event planner              & -    \\
    childcare worker           & -    \\ \bottomrule
    \end{tabular}
    \caption{\small {\bf Entities for scenario 2.} We choose seven most diverse professions and five least diverse professions, where the diversity is measured by Stable Diffusion 1.4 (SD1.4), SD2, and Dall-E 2.
    We also include five low-diversity professions for SD1.4 whose actual population includes more than 80\% women, with SD1.4 exacerbating gender stereotypes, whereas these professions show higher diversities in Dall-E 2 by over-representing male clusters \citep{luccioni2024stable_bias}.}
\label{appendix:tab:scenario_two_diversity}
\end{table}

\section{More experiments}

\subsection{More experimental results for Karlo UnCLIP decoder}
\label{appendix:sec:more_experiment_unclip_decoder}

\begin{figure}[h]
    \centering
    \includegraphics[width=\linewidth]{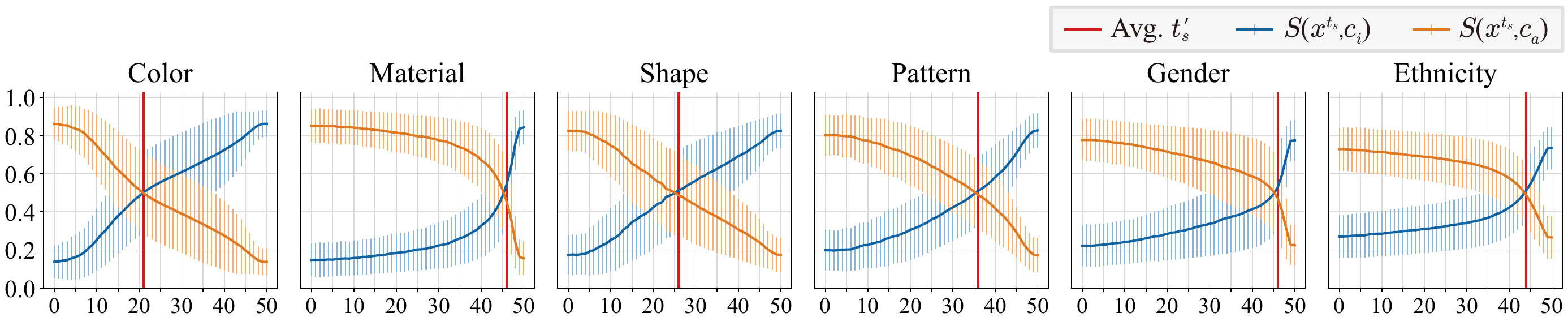}
    \caption{\small {\bf Switching point for the Karlo UnCLIP decoder.}}
    \label{appendix:fig:unclip_decoder1}
\end{figure}

\begin{figure}[h]
    \centering
    \includegraphics[width=\linewidth]{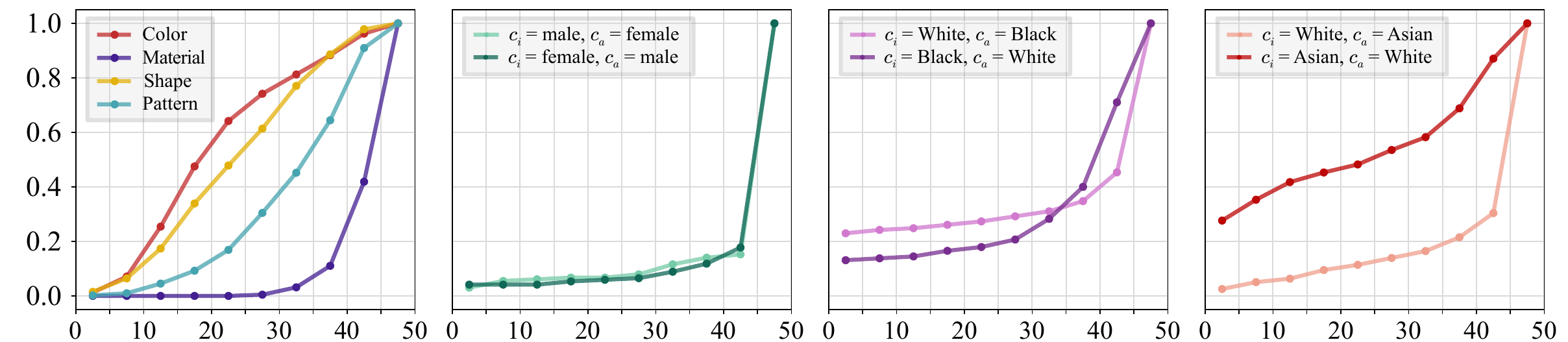}
    \caption{\small {\bf Cumulative histogram of the sample-wise switching timing for the Karlo UnCLIP decoder.}}
    \label{appendix:fig:unclip_decoder2}
\end{figure}

As we described in \cref{appendix:sec:diffusion_model_details}, the Karlo UnCLIP model consists of two parts and we only control the text condition of the prior module. In this subsection, we show the results when we generate an image latent by the prior module with text condition $c_i$, and then generate the pixel-level image by the decoder module, with the alternation of the prompt as proposed in our experiments.
\cref{appendix:fig:unclip_decoder1} and \ref{appendix:fig:unclip_decoder2} shows the results corresponding to the main results in the paper (\ie, \cref{fig:exp_main_switch} and \cref{fig:exp_main_human}). We observe that a similar early determination somewhat happens in this scenario, even if the given latent is already translated by the initial prompt $c_i$.


\subsection{Example of generated images}

\begin{figure}[h]
    \centering
    \includegraphics[width=\linewidth]{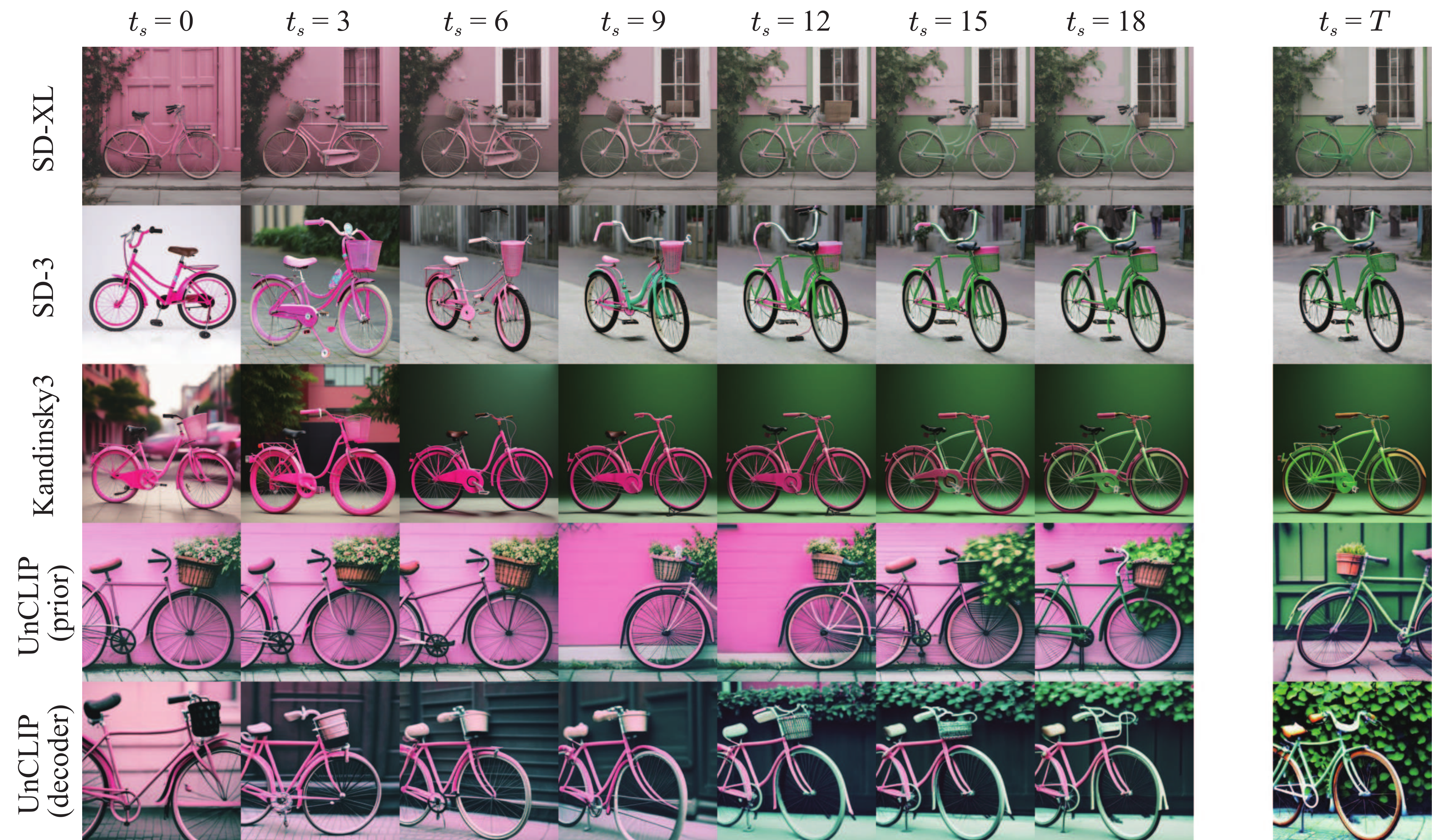}
    
    \small{$c_i$: "A photo of a green bicycle.", $c_a$: "A photo of a pink bicycle.".}
    
    \centering
    \includegraphics[width=\linewidth]{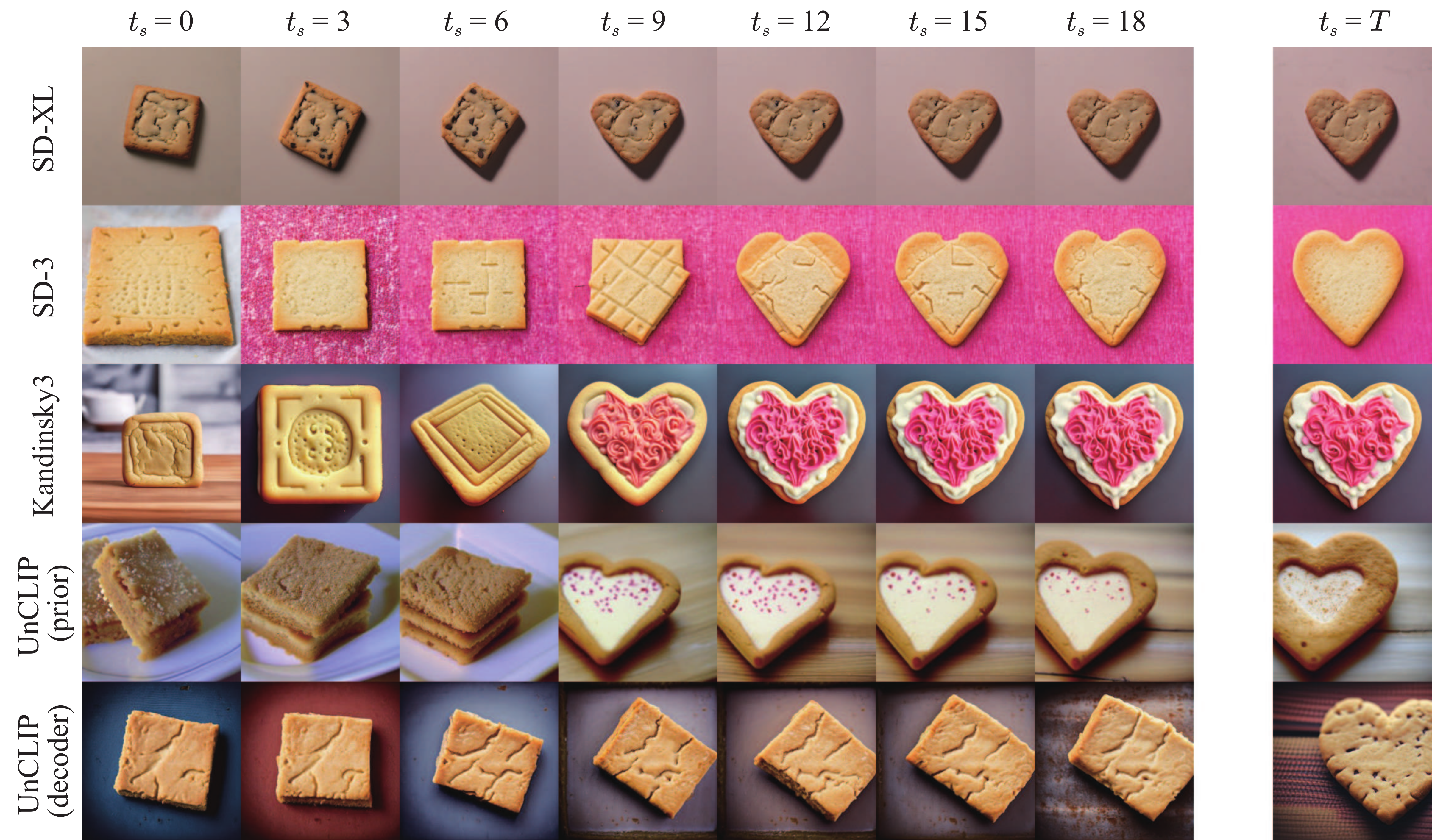}
    
    \small{$c_i$: "A photo of a heart-shaped cookie.", $c_a$: "A photo of a square cookie.".}
    \caption{\small {\bf Generated samples.}}
    \label{appendix:fig:generated_samples1}
\end{figure}

\begin{figure}[h]
    \centering
    \includegraphics[width=\linewidth]{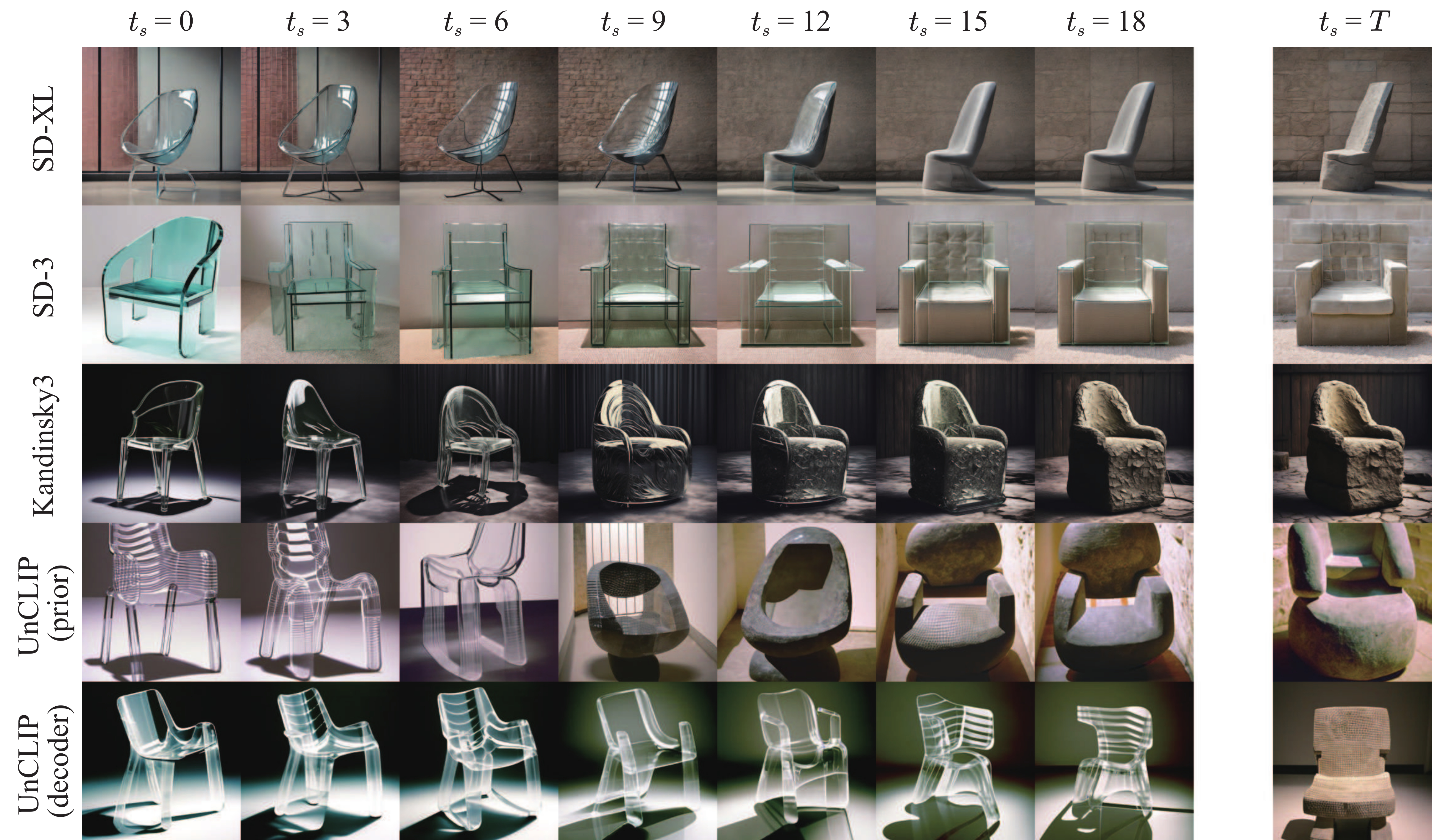}
    
    \small{$c_i$: "A photo of a stone chair.", $c_a$: "A photo of a glass chair.".}
    
    \centering
    \includegraphics[width=\linewidth]{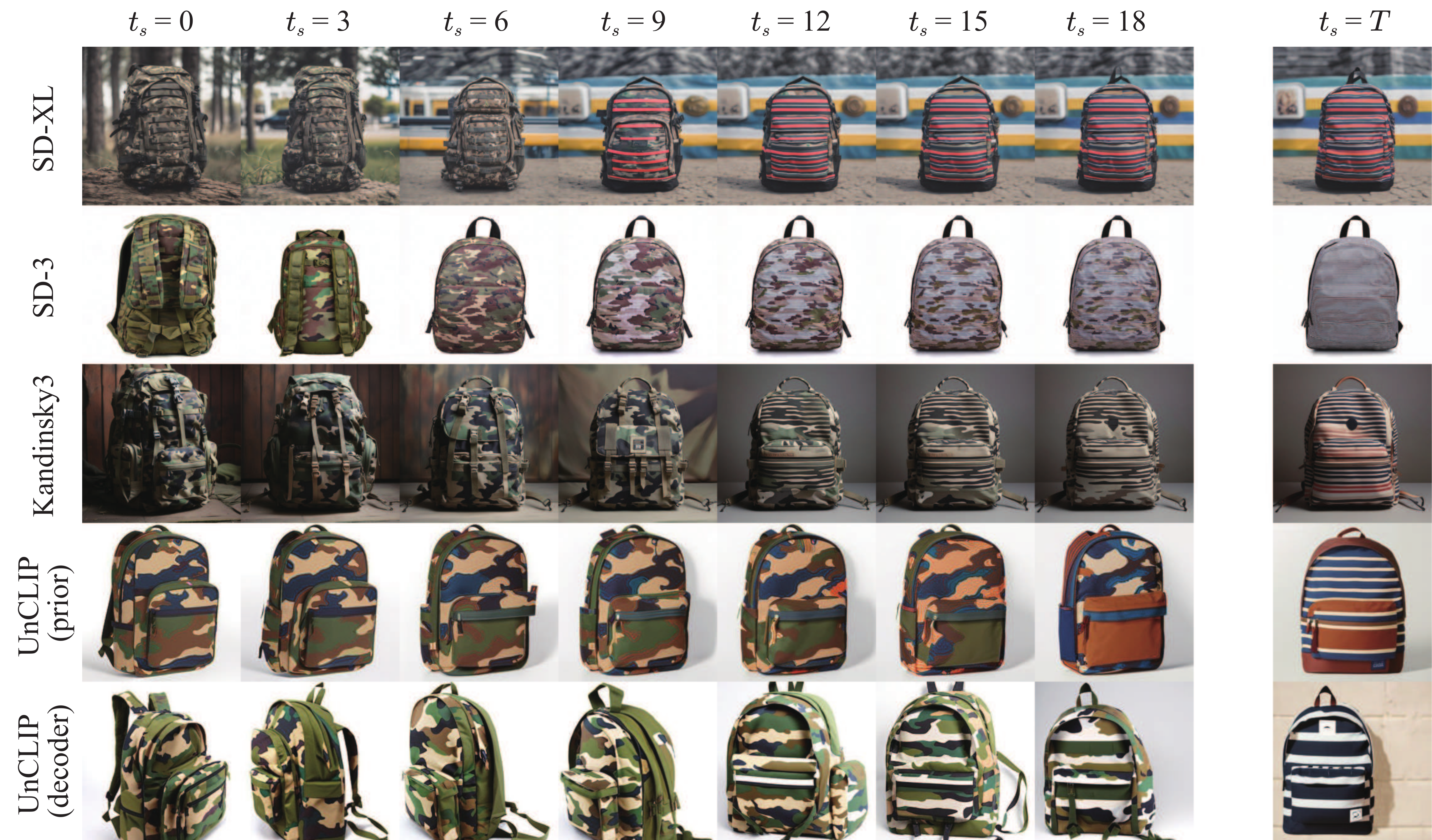}
    
    \small{$c_i$: "A photo of a stripes backpack", $c_a$: "A photo of a camouflage backpack.".}
    \caption{\small {\bf Generated samples.}}
    \label{appendix:fig:generated_samples2}
\end{figure}

\begin{figure}[h]
    \centering
    \includegraphics[width=\linewidth]{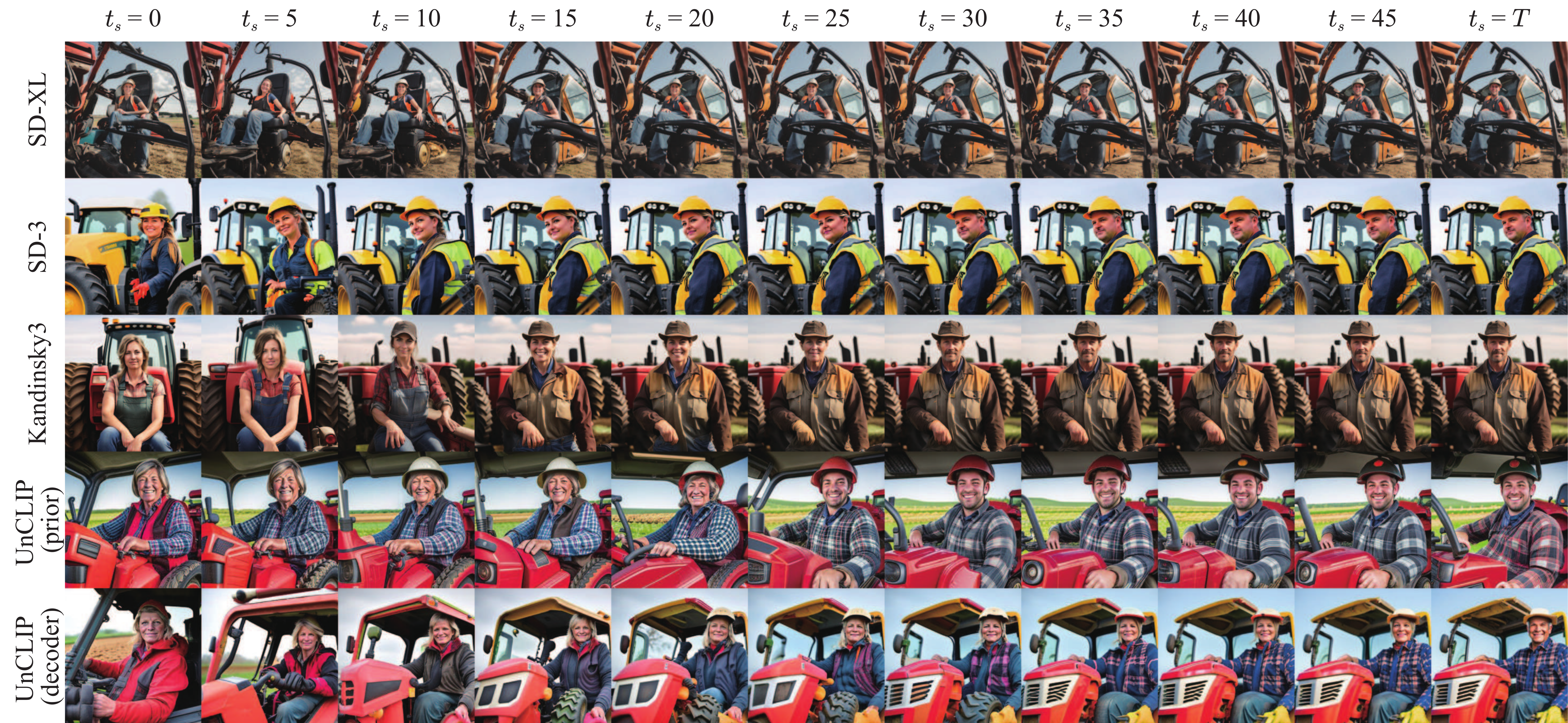}
    
    \small{$c_i$: "A photo of a male tractor operator.", $c_a$: "A photo of a female tractor operator.".}
    
    \centering
    \includegraphics[width=\linewidth]{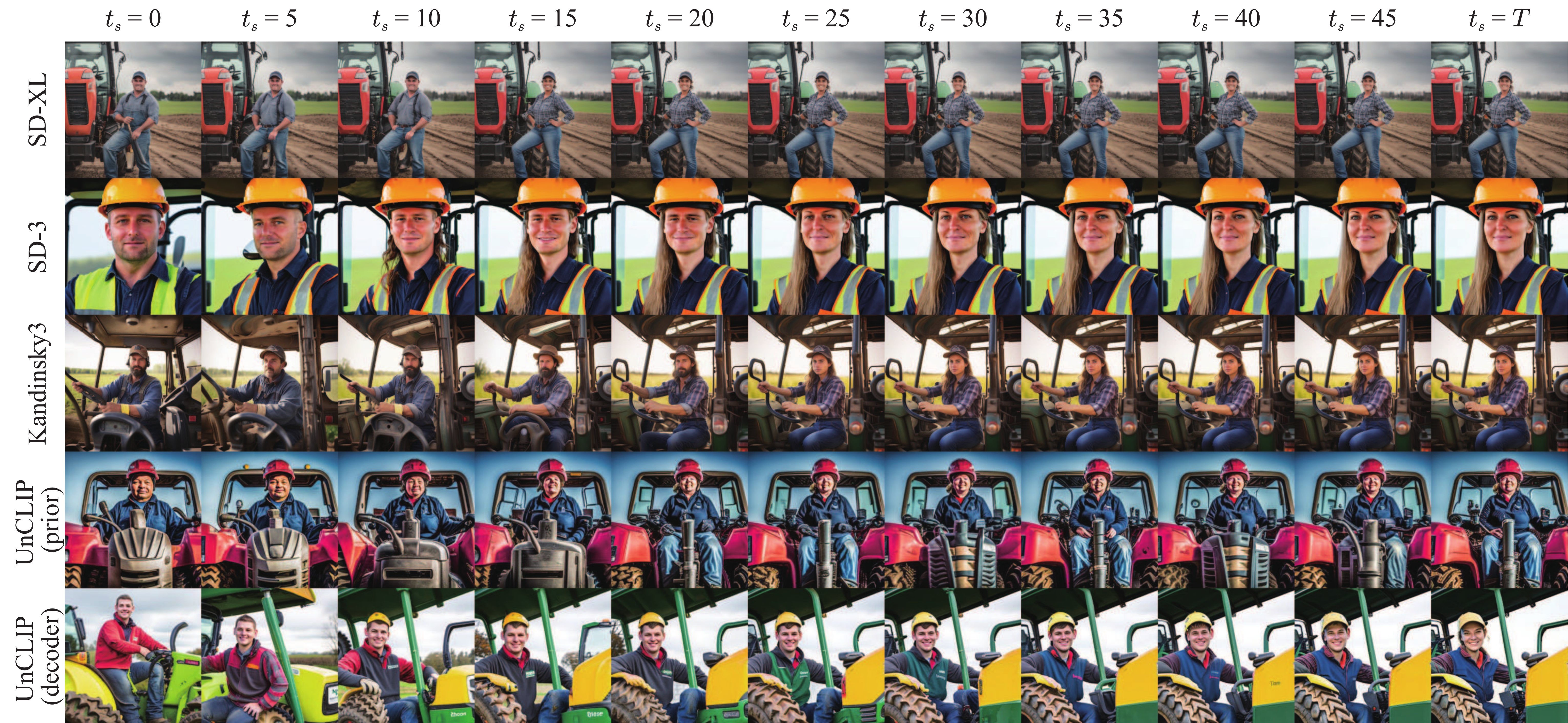}
    
    \small{$c_i$: "A photo of a female tractor operator.", $c_a$: "A photo of a male tractor operator.".}
    \caption{\small {\bf Generated samples.}}
    \label{appendix:fig:generated_samples3}
\end{figure}

\begin{figure}[h]
    \centering
    \includegraphics[width=\linewidth]{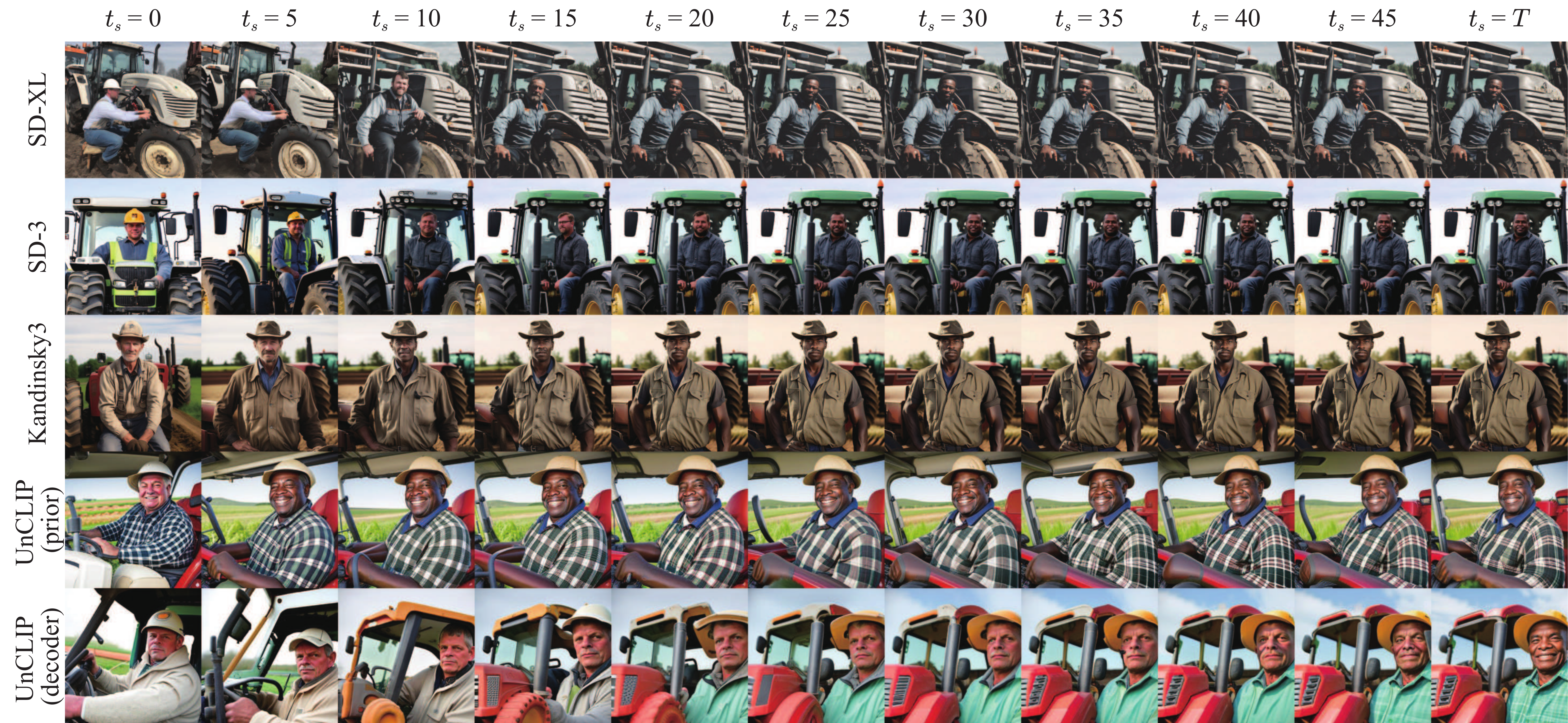}
    
    \small{$c_i$: "A photo of a Black tractor operator.", $c_a$: "A photo of a White tractor operator.".}
    
    \centering
    \includegraphics[width=\linewidth]{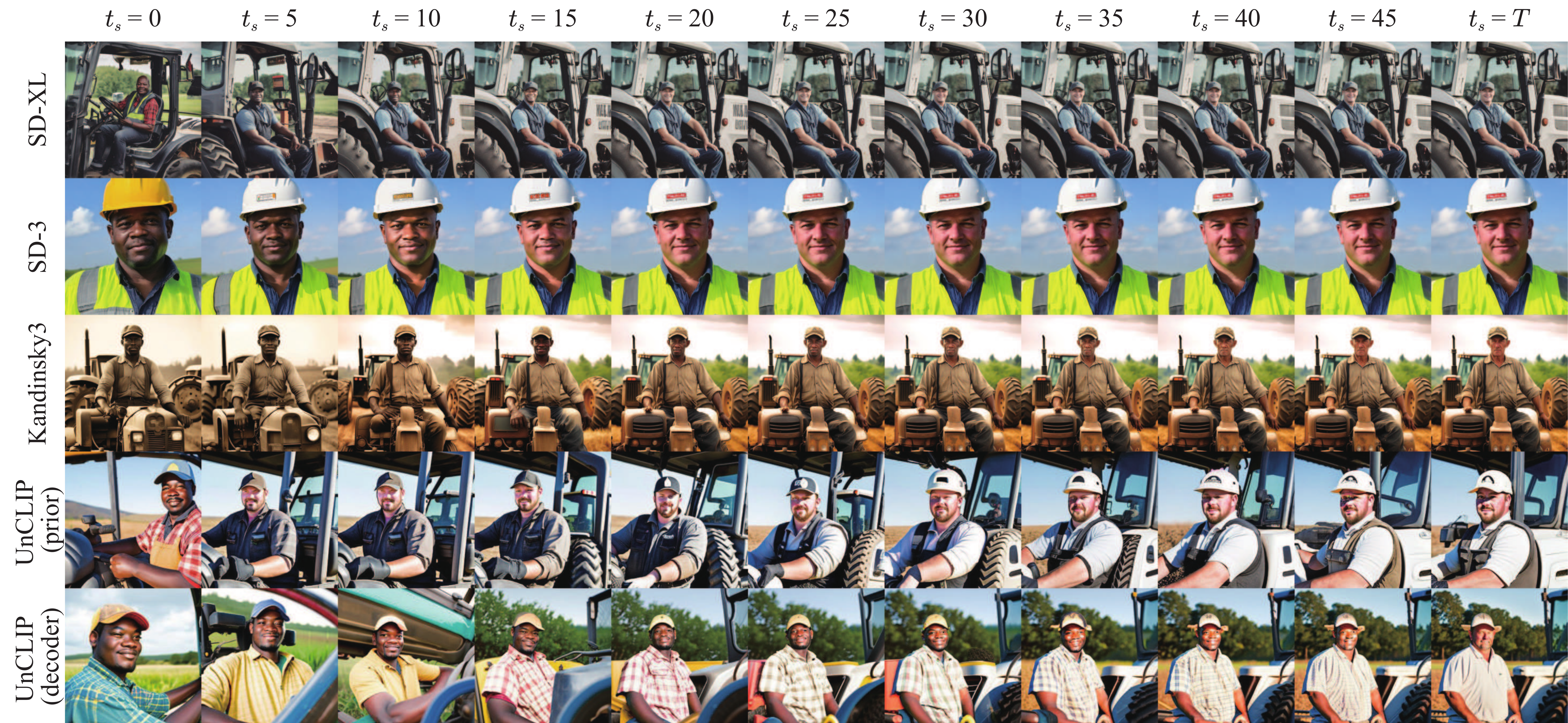}
    
    \small{$c_i$: "A photo of a White tractor operator.", $c_a$: "A photo of a Black tractor operator.".}
    \caption{\small {\bf Generated samples.}}
    \label{appendix:fig:generated_samples4}
\end{figure}

\begin{figure}[h]
    \centering
    \includegraphics[width=\linewidth]{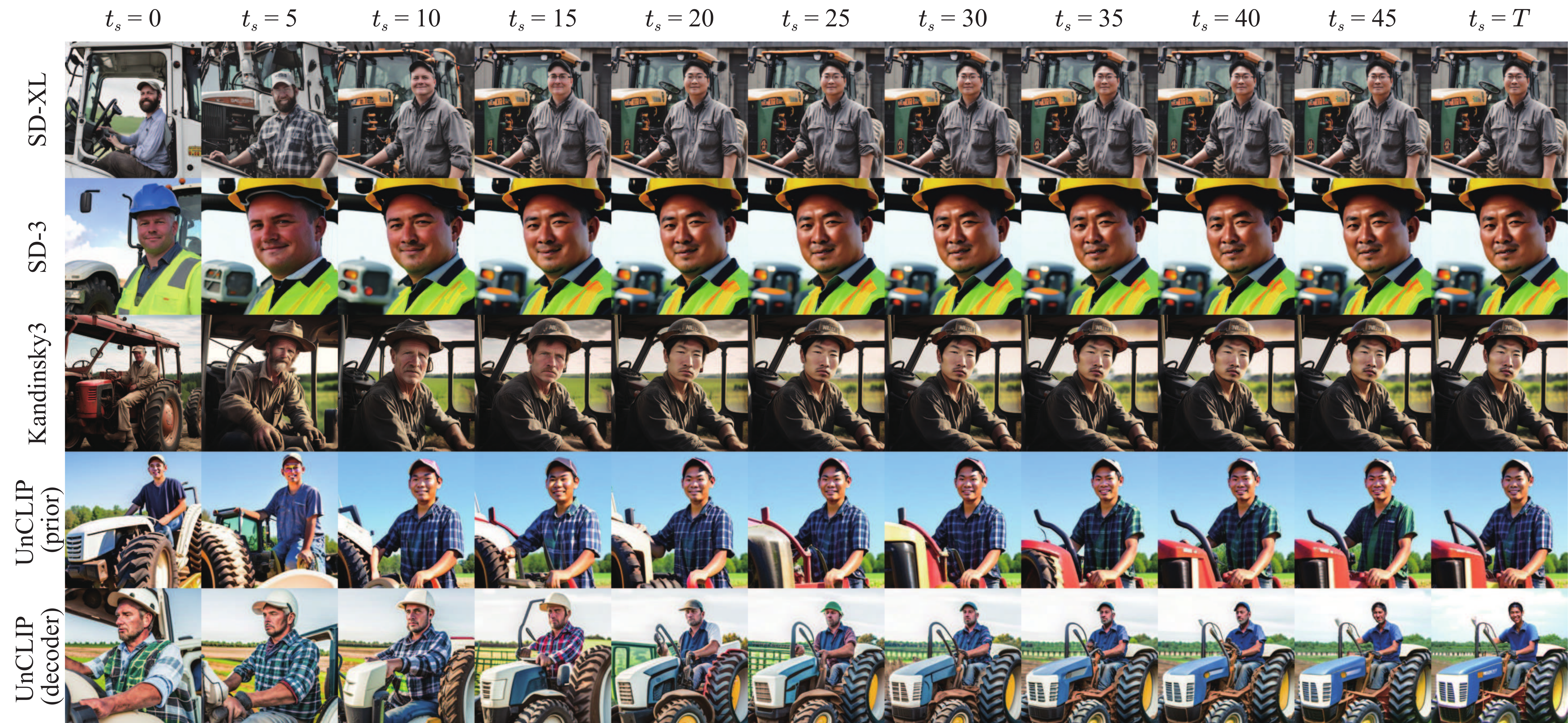}
    
    \small{$c_i$: "A photo of a Asian tractor operator.", $c_a$: "A photo of a White tractor operator.".}
    
    \centering
    \includegraphics[width=\linewidth]{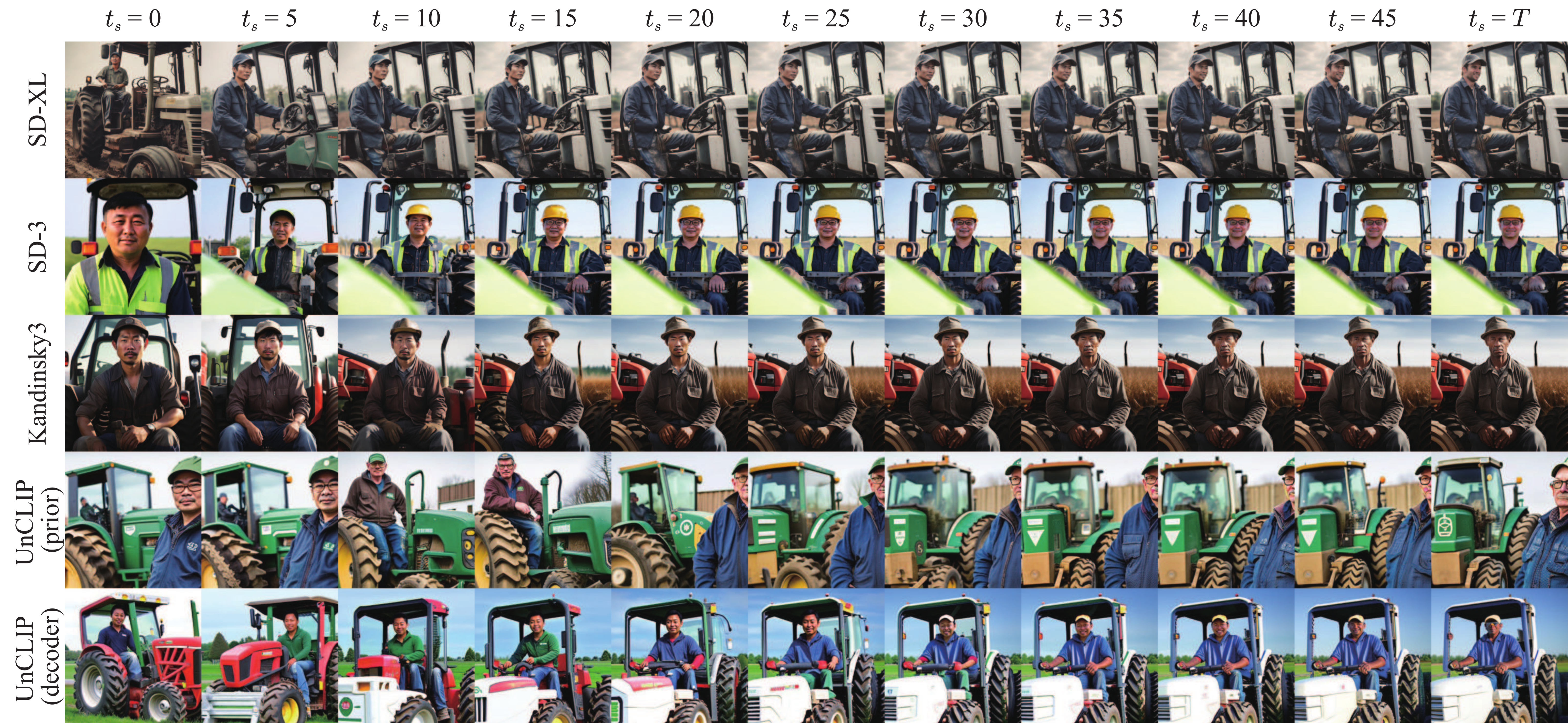}
    
    \small{$c_i$: "A photo of a White tractor operator.", $c_a$: "A photo of a Asian tractor operator.".}
    \caption{\small {\bf Generated samples.}}
    \label{appendix:fig:generated_samples5}
\end{figure}

We illustrate the generated images by altering the prompt from $c_i$ to $c_a$ in \cref{appendix:fig:generated_samples1},\ref{appendix:fig:generated_samples2}, \ref{appendix:fig:generated_samples3}, \ref{appendix:fig:generated_samples4} and \ref{appendix:fig:generated_samples5}. We can observe that the generated images progressively change their appearance reflecting the prompt $c_i$ or $c_a$.

\subsection{Progressive diffusion steps by difficulty}
\label{appendix:subsec:progressive_prompt}

In \cref{appendix:fig:progressive_prompt1}, \ref{appendix:fig:progressive_prompt2}, and \ref{appendix:fig:progressive_prompt3}, we show the examples when our progressive prompt altering helps to generate complex and difficult prompts.
Here we use complex prompts with two distinct features. For example, ``a photo of gray zigzag jacket'' (\cref{appendix:fig:progressive_prompt1}) contains ``gray'' and ``zigzag'' attributes. We start with ``a photo of jacket'' and add ``gray'' and ``zigzag'' at different timestamps.
For example, we alter ``a photo of jacket'' to ``a photo of gray jacket'' and then we alter again the prompt to ``a photo of gray zigzag jacket''.
The figures show the generated images for different altering timestamps for each attribute. Interestingly, while the generated images only guided by the full prompt often fail to generate the desired attribute. For example, \cref{appendix:fig:progressive_prompt3} shows that the generated images guided by the original prompt (\ie, the most right below images) fail to capture both color and shape. On the other hand, when we control the altering timing, the generated images can capture both features without ignoring any of the features.

\begin{figure*}[ht!]
    \centering
    \includegraphics[width=0.63\linewidth]{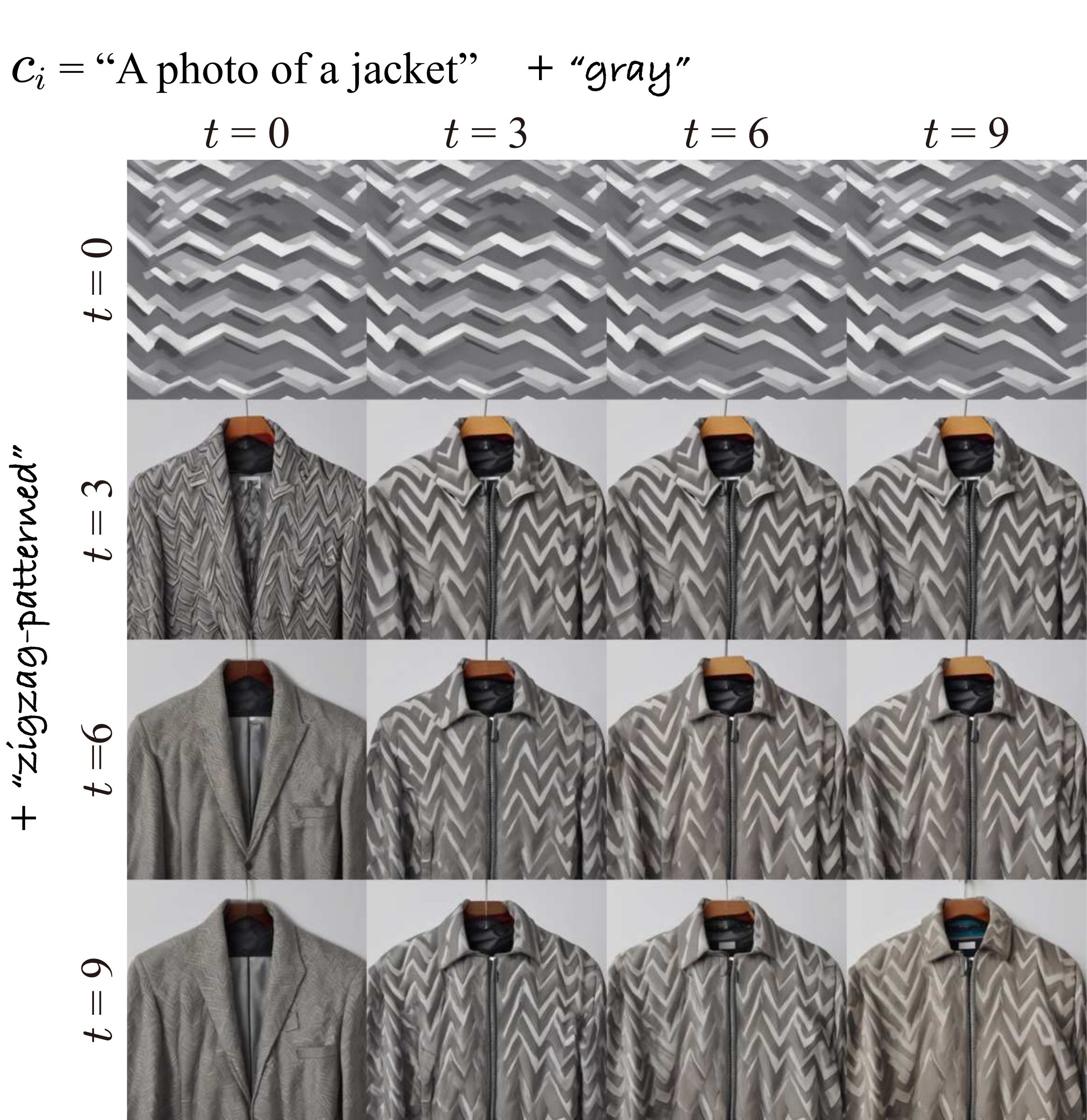}

    \includegraphics[width=0.63\linewidth]{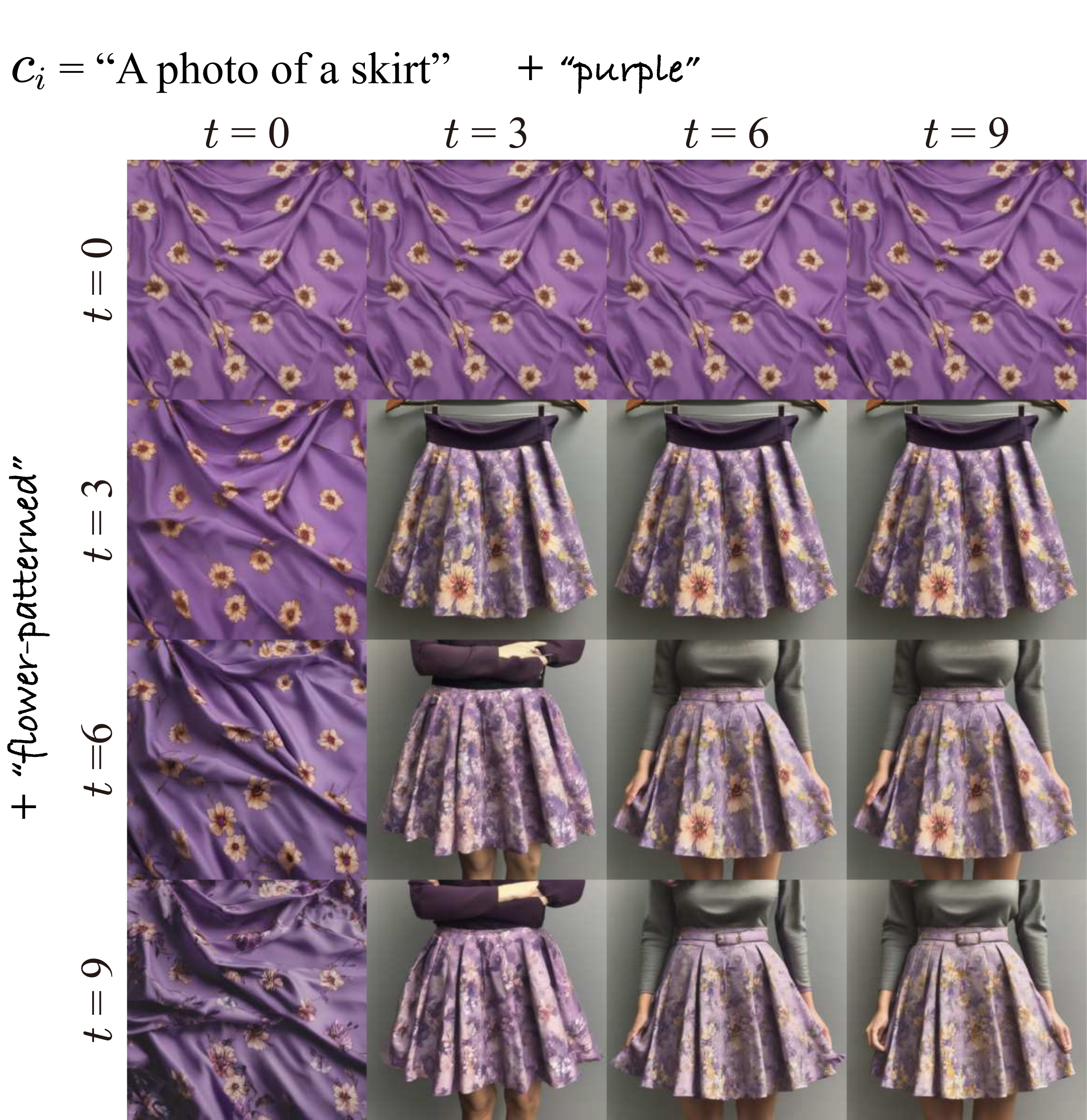}

    \caption{\small {\bf Generated images with progressive diffusion steps (pattern).} The most left top image is only guided by $c_i$ and the most right bottom image is only guided by $c_a$.}
    \label{appendix:fig:progressive_prompt1}
\end{figure*}

\begin{figure*}[ht!]
    \centering
    \includegraphics[width=0.63\linewidth]{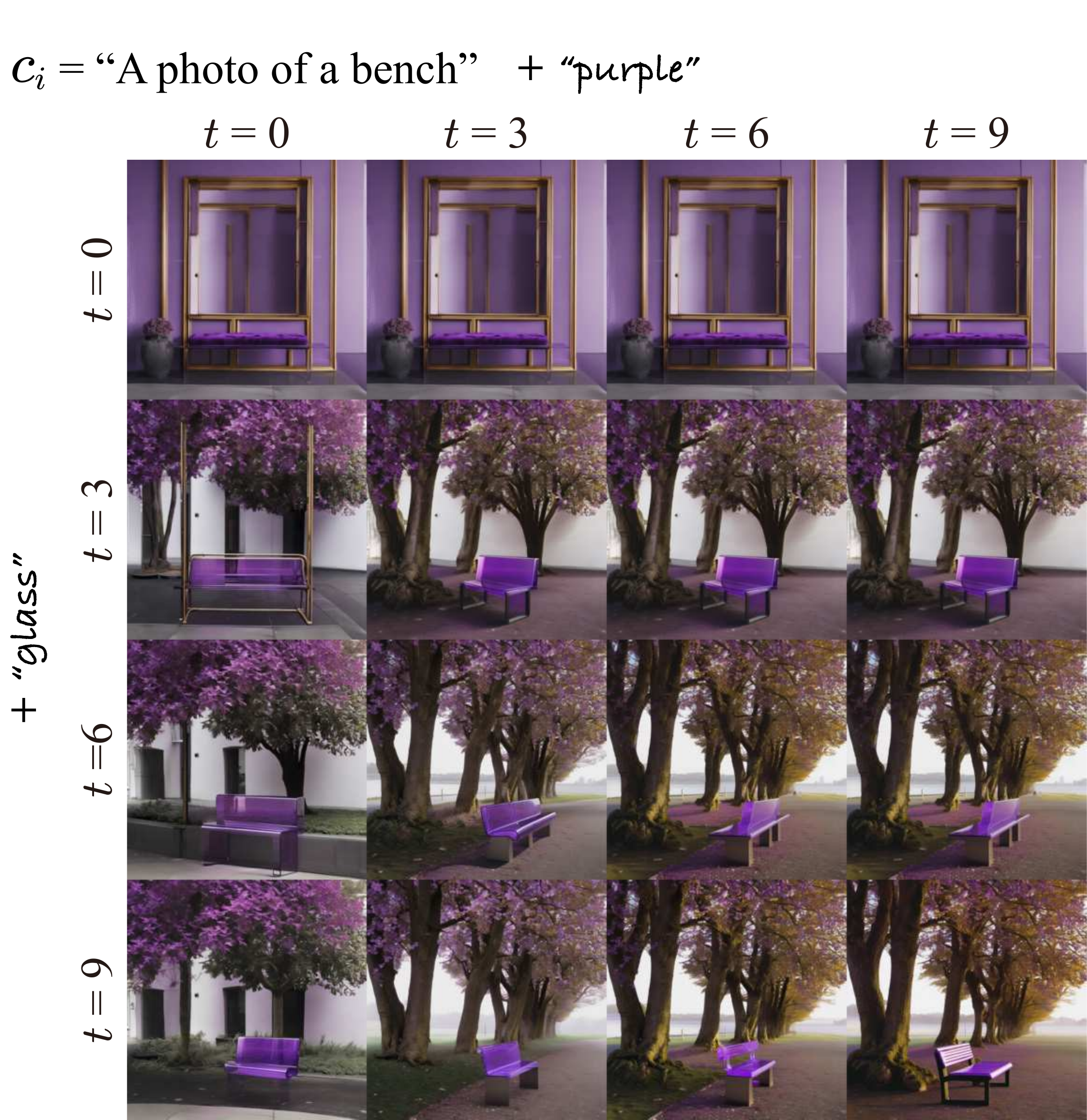}

    \includegraphics[width=0.63\linewidth]{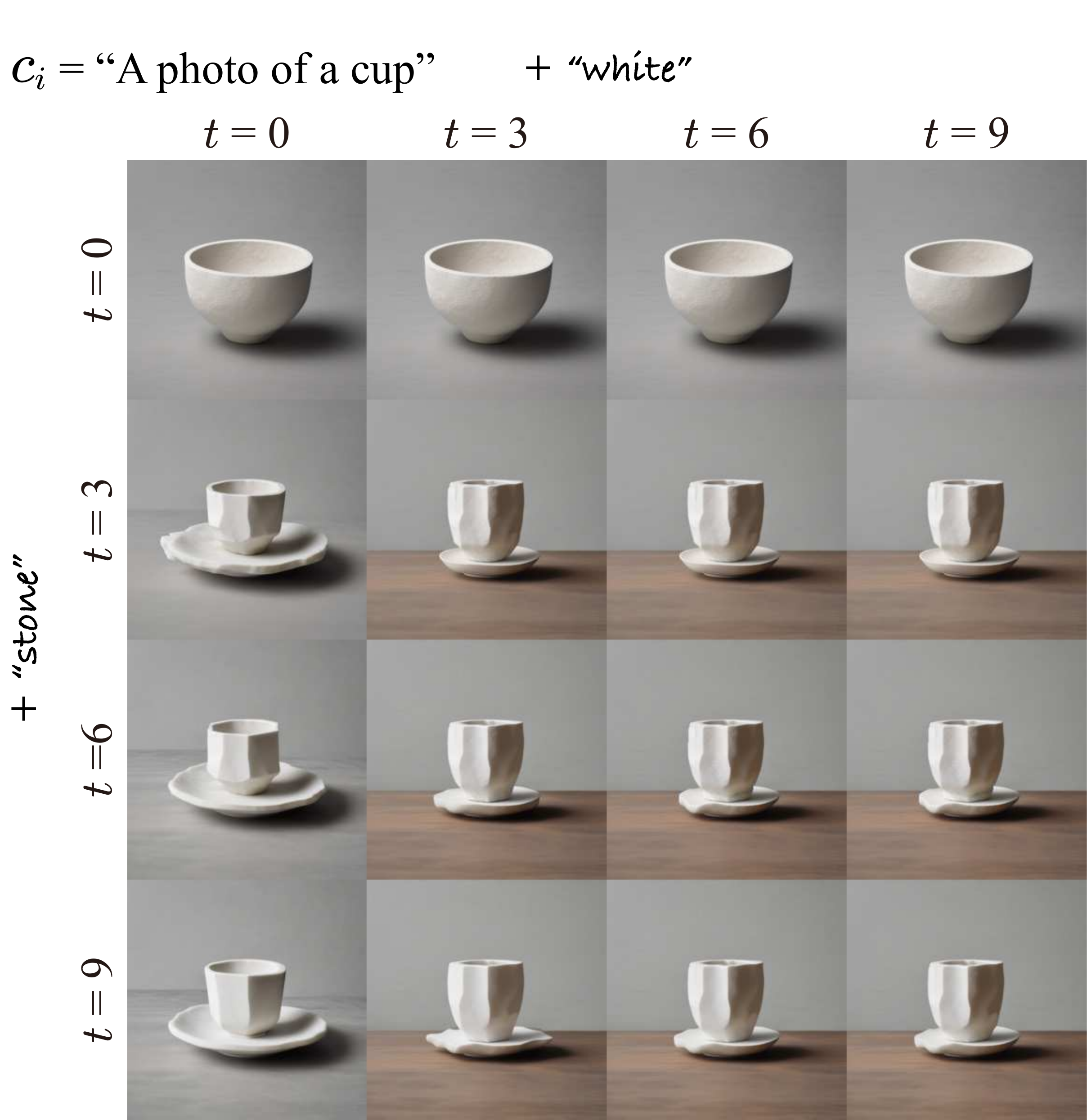}

    \caption{\small {\bf Generated images with progressive diffusion steps (material).} The details are the same as \cref{appendix:fig:progressive_prompt1}.}
    \label{appendix:fig:progressive_prompt2}
\end{figure*}

\begin{figure*}[ht!]
    \centering
    \includegraphics[width=0.63\linewidth]{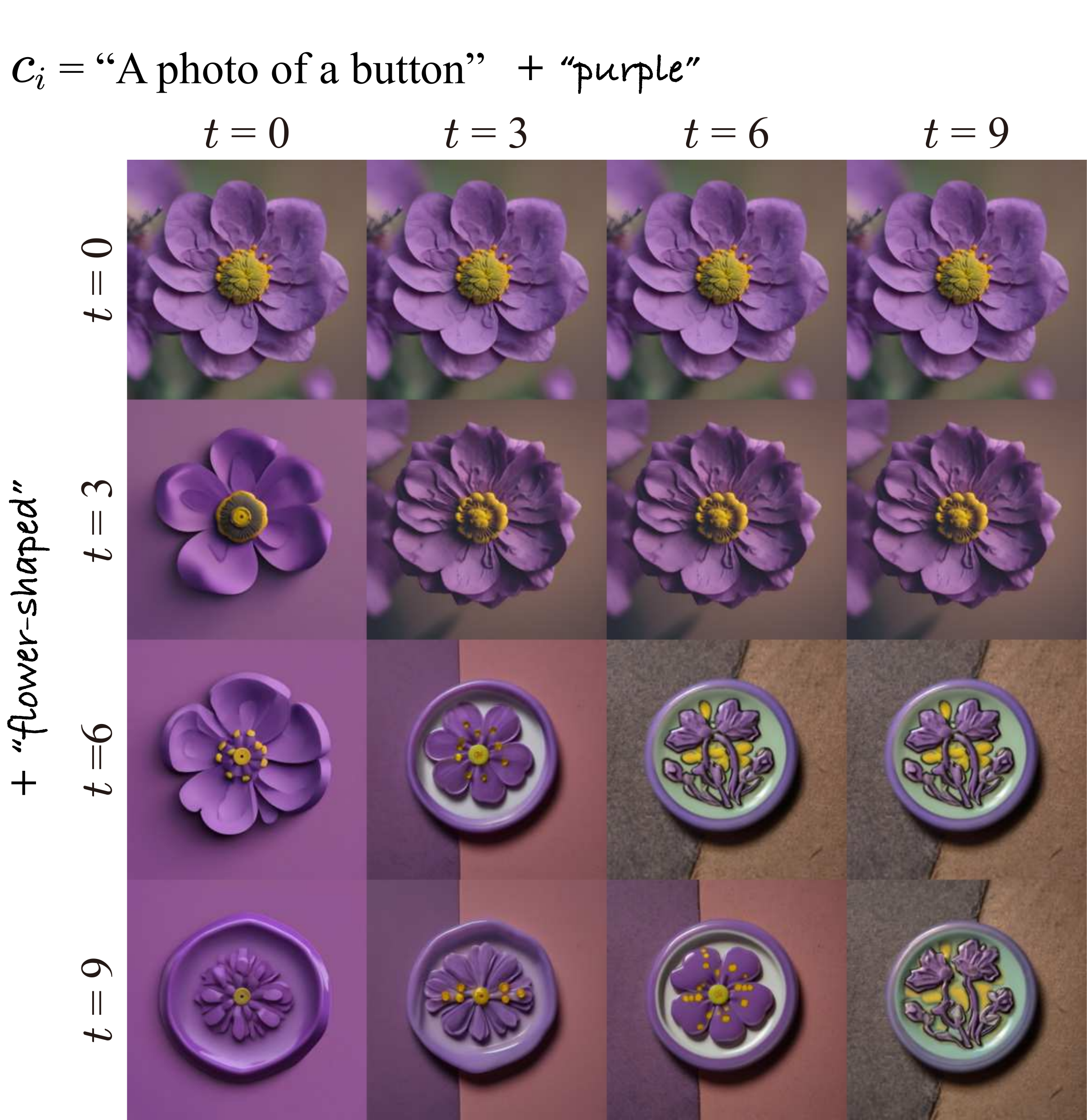}

    \includegraphics[width=0.63\linewidth]{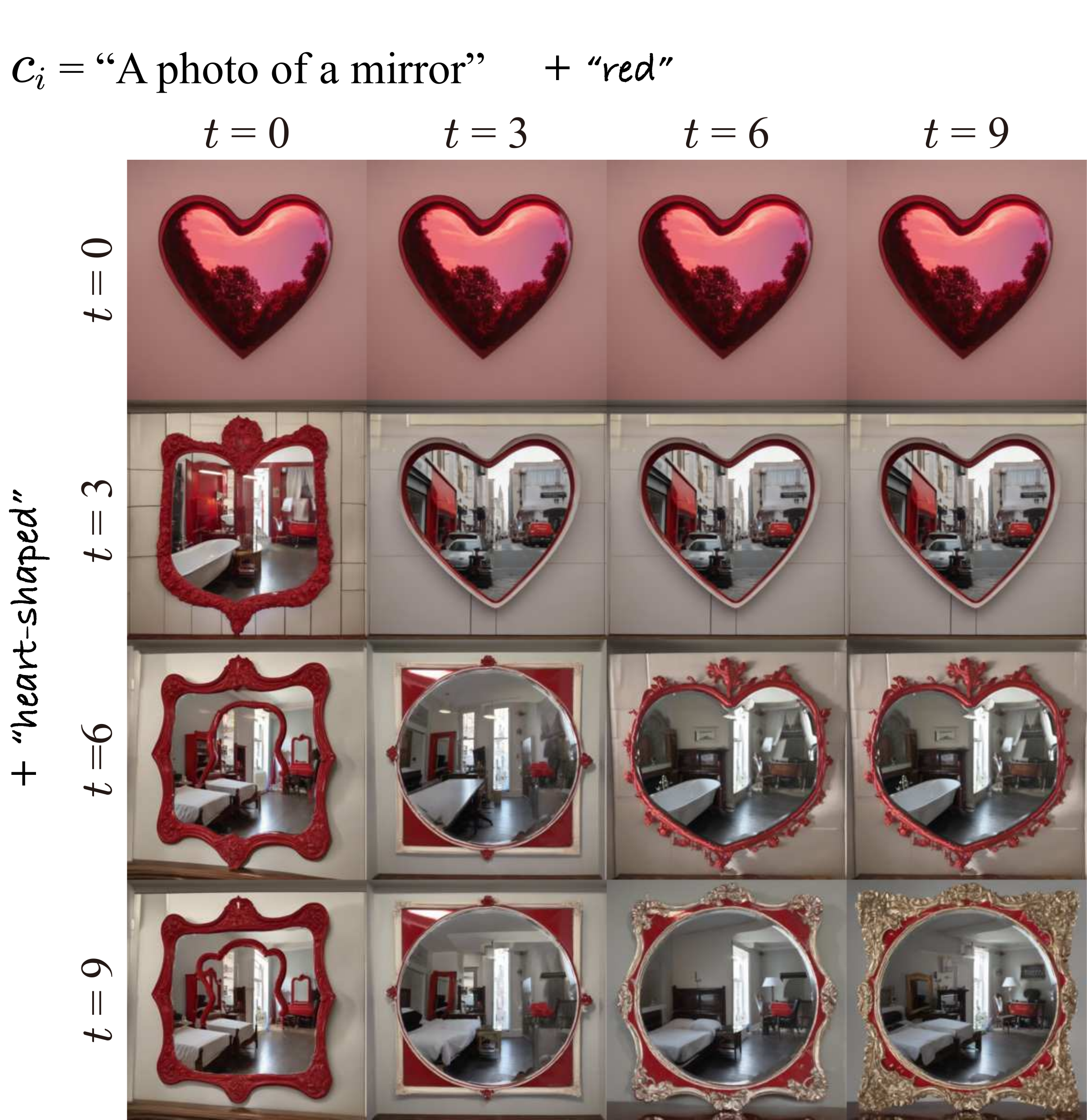}

    \caption{\small {\bf Generated images with progressive diffusion steps (shape).} The details are the same as \cref{appendix:fig:progressive_prompt1}.}
    \label{appendix:fig:progressive_prompt3}
\end{figure*}

\end{document}